\documentclass[11pt]{article}

\usepackage{acl}

\usepackage{times}
\usepackage{latexsym}

\usepackage[T1]{fontenc}

\usepackage[utf8]{inputenc}

\usepackage{microtype}

\title{\textsc{Hatemoji}: A Test Suite and Adversarially-Generated Dataset for Benchmarking and Detecting Emoji-Based Hate}

\author{Hannah Rose Kirk$^{1,2\ddagger}$, Bertie Vidgen$^{1,2}$, Paul Röttger$^{1,2}$, Tristan Thrush$^{3}$\thanks{~~TT's work was done at Facebook AI Research.}, Scott A. Hale$^{1, 2, 4}$\\
  { $^1$University of Oxford, $^2$The Alan Turing Institute, $^3$Hugging Face, $^4$Meedan} \\
  { $^\ddagger$hannah.kirk@oii.ox.ac.uk}\\
}

\usepackage{cleveref}
\crefname{section}{Sec.}{Sec.}
\crefname{Section}{Sec.}{Sec.}
\crefname{table}{Tab.}{Tab.}
\crefname{appendix_table}{Tab.}{Tab.}
\crefname{Table}{Tab.}{Tab.}
\crefname{Figure}{Fig.}{Fig.}
\crefname{figure}{Fig.}{Fig.}
\crefname{appendix}{Appendix}{Appendix}
\crefname{chapter}{Chapter}{Chapter}

\usepackage[whatsapp]{emoji}

\usepackage{booktabs}
\newlength{\origaboverulesep}
\newlength{\origbelowrulesep}
\usepackage{marginnote}
\usepackage{fancyhdr}

\fancypagestyle{firstpage}
{
    \fancyhead[L]{}    
    \fancyhead[R]{This paper is accepted for publication at NAACL 2022. The official version of record is in the ACL Anthology.}
}

\newcommand\Tstrut{}%
\usepackage{colortbl}
\usepackage{amssymb}
\usepackage{multirow}
\usepackage{hhline}
\usepackage[T1]{fontenc}
\usepackage{makecell}
\usepackage{multirow, makecell}
\usepackage{rotating}
\usepackage{arydshln}
\usepackage{longtable}
\usepackage{bm}
\usepackage{float}

\usepackage{soul}
\usepackage{color}

\definecolor{hate}{rgb}{0.98, 0.81, 0.69}
\definecolor{nothate}{rgb}{0.67, 0.9, 0.93}
\DeclareRobustCommand{\hlcyan}[1]{{\sethlcolor{nothate}\hl{#1}}}
\DeclareRobustCommand{\hlorg}[1]{{\sethlcolor{hate}\hl{#1}}}
\usepackage[T1]{fontenc}

\date{}

\usepackage{fancyhdr}
\begin{document}
\maketitle
\thispagestyle{firstpage}
\begin{abstract}
Detecting online hate is a complex task, and low-performing models have harmful consequences when used for sensitive applications such as content moderation.
Emoji-based hate is an emerging challenge for automated detection.
We present \textsc{HatemojiCheck}, a test suite of $3{,}930$ short-form statements that allows us to evaluate performance on hateful language expressed with emoji.
Using the test suite, we expose weaknesses in existing hate detection models.
To address these weaknesses, we create the \textsc{HatemojiBuild} dataset using a human-and-model-in-the-loop approach. Models built with these $5{,}912$ adversarial examples perform substantially better at detecting emoji-based hate, while retaining strong performance on text-only hate. Both \textsc{HatemojiCheck} and \textsc{HatemojiBuild} are made publicly available.\footnote{See our \href{https://github.com/HannahKirk/Hatemoji}{Github Repository}. \href{https://huggingface.co/datasets/HannahRoseKirk/HatemojiCheck}{\textsc{HatemojiCheck}}, \href{https://huggingface.co/datasets/HannahRoseKirk/HatemojiBuild}{\textsc{HatemojiBuild}} and the final \href{https://huggingface.co/HannahRoseKirk/Hatemoji}{\textsc{Hatemoji} Model} are also available on HuggingFace.}

\end{abstract}

\vspace{-2em}
\textcolor{red}{\paragraph{Content Warning} This article contains examples of hateful language from \textsc{HatemojiCheck} to illustrate its composition.
Examples are quoted verbatim, except for slurs and profanity in text, for which the first vowel is replaced with an asterisk. The authors oppose the use of hateful language.}

\section{Introduction}
Online hate harms its targets, disrupts online communities, pollutes civic discourse and reinforces social power imbalances \citep{gelberEvidencingHarms2016}.
The sheer scale of hateful content online has led to widespread use of automated detection systems to find, monitor and stop it \citep{ vidgenChallengesFrontiers2019a, gillespieContentModeration2020a}.
However, hateful content is complex and diverse, which makes it challenging to detection systems.
One particular challenge is the use of emoji for expressing hate.
Emoji are pictorial representations that can be embedded in text, allowing complex emotions, actions and intentions to be displayed concisely \citep{rodriguesLisbonEmoji2018}. Over 95\% of internet users have used an emoji and 10 million are sent every day \citep{brandwatchEmojiReport2018}.
Following England's defeat in the Euro 2020 football final, there was widespread racist use of emoji such as \emoji{monkey2}, \emoji{banana} and \emoji{watermelon} \citep{jamiesonRacistComments2020}.
This paper focuses on emoji-based hate, answering two research questions: 
\begin{description}
    \item[\textbf{RQ1}] What are the weaknesses of current detection systems for hate expressed with emoji? 
    \item[\textbf{RQ2}] To what extent can human-and-model-in-the-loop training improve the performance of detection systems for emoji-based hate?
\end{description}

To answer RQ1, we present \textsc{HatemojiCheck}, a suite of functional tests for emoji-based hate.
We provide $3{,}930$ test cases for seven functionalities, covering six identities. $2{,}126$ original test cases are matched with three types of challenging perturbations to enable accurate evaluation of model decision boundaries \citep{gardnerEvaluatingModels2020a}. 
We use \textsc{HatemojiCheck} to assess Google Jigsaw's Perspective API as well as models trained on academic datasets, exposing critical model weaknesses.

To answer RQ2 and address the model weaknesses identified by \textsc{HatemojiCheck}, we implement a human-and-model-in-the-loop dynamic training scheme.
We build on the work of \citet{vidgenLearningWorst2021}, who used this approach for textual hate.
Our work begins where their study ends. We conduct three rounds of adversarial data generation focused explicitly on emoji-based hate, tasking annotators to generate sentences that the model-in-the-loop misclassifies.
This process yields a dataset of $5{,}912$ entries, half of which are challenging contrasts. We call this dataset \textsc{HatemojiBuild}.
The dataset is evenly split between hate and non-hate, and each hateful entry has labels for the type and target. 
Between each round, the model-in-the-loop is re-trained so that annotators are trying to trick a progressively stronger and more `emoji-aware' model. 
Relative to existing commercial and academic models, our models improve performance on the detection of emoji-based hate, without sacrificing performance on text-only hate. 

We make several contributions: (1) we construct \textsc{HatemojiCheck}, which tests key types of emoji-based hate as separate functionalities, (2) we evaluate the performance of existing academic and commercial models at detecting emoji-based hate, (3) we present \textsc{HatemojiBuild}, a labeled emoji-based hate speech dataset that is adversarially-generated for model training and (4) we train models that can accurately detect emoji-based hate. These data-centric contributions demonstrate the benefits of systematic and granular evaluation, and the need to diversify how hate detection systems are trained.
We make both \textsc{HatemojiCheck} and \textsc{HatemojiBuild} publicly available.

\paragraph{Definition of Hate} We use the United Nations definition of hate speech: ``any kind of communication in speech, writing or behavior, that attacks or uses pejorative or discriminatory language with reference to a person or a group on the basis of who they are, in other words, based on their religion, ethnicity, nationality, race, color, descent, gender or other identity factor'' \citep[p.2]{unitednationsStrategyPlan2019}.\footnote{We recognize the presence of annotator bias in cultural interpretations of this definition, and provide data statements for both datasets in the appendix.}

\section{\textsc{HatemojiCheck}: Functional Tests for Emoji-Based Hate}
The concept of functional testing, an expected input-output behavior \citep{beizerBlackboxTesting1995}, has been adapted from software engineering for NLP tasks \citep{ribeiroAccuracyBehavioral2020}, including hate speech detection \citep{rottgerHateCheckFunctional2021}. These diagnostic tests identify model vulnerabilities to intentionally simple and non-ambiguous constructions of hate so comprise a minimal performance standard. 

\subsection{Identifying Functionalities}
A functionality describes the ability of a model to provide a correct classification when presented with hateful or non-hateful content \citep{rottgerHateCheckFunctional2021}. 
Each functionality has a set of corresponding test cases that share one gold-standard label.
We select the functionalities in \textsc{HatemojiCheck} to be
(1) realistic: they capture real-world uses of emoji-based hate,
(2) unique: each covers a distinct aspect of emoji-based hate, without overlaps between functionalities, and
(3) unambiguous: they have clear gold-standard labels.
The functionalities are motivated from two perspectives:

\paragraph{Previous Literature} We identify distinct uses of emoji in online communications, particularly abuse. This includes appending emoji to the end of otherwise neutral statements to reveal emotive intent \citep{wiegandExploitingEmojis2021}, using emoji as a `pictorial slur' to attack or dehumanize groups \citep{bickAnnotatingEmoticons2020, wiegandExploitingEmojis2021} and substituting emoji for identity terms or threatening verbs \citep{samghabadiAttendingEmotions2019, bickAnnotatingEmoticons2020, wiegandExploitingEmojis2021}.
We also review a range of literature on online hate, identifying distinct types of hateful language \citep{talatUnderstandingAbuse2017, zampieriSemEval2019Task2019, vidgenIntroducingCAD2021}.

\paragraph{Empirical Investigation} Based on the literature, we define a list of potentially hateful emoji and words, and use Twitter's Streaming API to search for the Cartesian products of emoji--emoji and emoji--word pairs over a two week period. To identify different forms of emoji-based hate, we apply a grounded theory approach \citep{corbinGroundedTheory1990} on a sample of $3{,}295$ tweets, splitting out distinctive categories, and recursively selecting sub-categories until all key parts of the data are captured and the framework is `saturated'.

\begin{table*}
\centering
\footnotesize
\caption{Summary of the seven functionalities in \textsc{HatemojiCheck}.}
\label{tab:hmc_summary}
\setlength{\tabcolsep}{1.8pt}
\resizebox{\textwidth}{!}{
\begin{tabular}{@{}lcccc@{}}
\toprule
\multicolumn{1}{c}{\textbf{Functionality (Emoji Used)}}                                                                                                                                                                                                                                                                 & \textbf{Templates} & \textbf{Emoji} & \textbf{Identities} & \textbf{Cases} \\ \midrule
\textbf{\textit{F1}} \textit{Verb Swap}: Swap threatening verb (\emoji{stab}, \emoji{bomb}, \emoji{punch}, \emoji{gun}, \emoji{skull})                                                                                                                                                                     & 10                           & 5                        & 6                             & 300                      \\
\textbf{\textit{F2}} \textit{Identity Swap}: Swap protected identity (\emoji{woman}, \emoji{trans}, \emoji{gay}, \emoji{black}, \emoji{dis}, \emoji{muslim})                                                                                                                                               & 20                           & 1                        & 6                             & 120                      \\
\textbf{\textit{F3}} \textit{Descriptor Swap}: Swap descriptor noun/emotion ( \emoji{rat}, \emoji{pig}, \emoji{shit}, \emoji{virus}, \emoji{mosquito}, \emoji{monkey}, \emoji{gorilla}, \emoji{goat}, \emoji{camel} / \emoji{vomit}, \emoji{sick_mask}, \emoji{green_face}) & 10                           & 3--7                     & 6                             & 260                      \\
\textbf{\textit{F4}} \textit{Double Swap}: Combine two of the above swaps                                                                                                                                                                                                                                  & 10                           & 3--7                     & 6                             & 288                      \\
\textbf{\textit{F5}} \textit{Append}: Append negative emoji to neutral text (\emoji{swear}, \emoji{red_angry}, \emoji{vomit}, \emoji{green_face}, \emoji{huff})                                                                                                                                                & 10                           & 3--5                     & 6                             & 288                      \\
\textbf{\textit{F6}} \textit{Positive Confounder}: Append positive emoji to hateful text (\emoji{heart}, \emoji{smiley}, \emoji{angel}, \emoji{love_eyes}, \emoji{praise})                                                                                                                                     & 13                           & 5                        & 6                             & 440                      \\
\textbf{\textit{F7}} \textit{Emoji Leetspeak}: Substitute for character or word-piece (\emoji{0}, \emoji{1}, \emoji{4}, \emoji{3}, \emoji{star}, \emoji{heart})                                                                                                                                            & 15                           & 3--4                      & 6 (+16 slurs)                 & 430                      \\ \hline
\multicolumn{1}{r}{\textbf{Totals of Unique Templates, Emoji, Identity and Test Cases}}                                                                                                                                                                                                                    & 88                           & 36                       & 6                             & 2,126                     \\ \bottomrule
\end{tabular}
}
\end{table*}
\subsection{Functionalities in \textsc{HatemojiCheck}}
\footnotesize{\centering \textcolor{red}{This section presents verbatim examples of emoji-based hate (in \textcolor{gray}{gray text}). For masked versions, see \cref{tab:hmc_summary} and \cref{tab:func_hmc}.}}
\normalsize
\textsc{HatemojiCheck} has seven functionalities (see \cref{tab:hmc_summary}). \noindent \textbf{F1 Verb Swap} tests threatening hate where the verb is replaced with the equivalent emoji: \emoji{gun}, \emoji{punch}, \emoji{bomb}, \emoji{stab}, \emoji{skull}.
It includes both direct threats and normative statements of threat. \textbf{F2 Identity Swap} tests derogatory hate where the identity term is replaced with an emoji representation: woman (\emoji{woman}), trans people (\emoji{trans}), gay people (\emoji{gay}), disabled people (\emoji{dis}), Black people (\emoji{black}) and Muslims (\emoji{muslim}).\footnote{ \emoji{muslim} was commonly used to denote Muslims on Twitter, despite its Unicode description being `person wearing a turban'.}
\textbf{F3 Descriptor Swap} tests hate where the negative descriptor is replaced with the equivalent emoji. It includes both dehumanizing language and expressions of disgust. For dehumanizing language, five descriptors are used: \emoji{rat}, \emoji{pig}, \emoji{shit}, \emoji{mosquito}, \emoji{virus}. Based on our Twitter analysis, target-specific descriptors are used for Black people: (\emoji{monkey}, \emoji{gorilla}) and Muslims: (\emoji{goat}, \emoji{camel}).
For expressions of disgust, three descriptors are used: \emoji{vomit}, \emoji{sick_mask}, \emoji{green_face}.
\textbf{F4 Double Swap} tests a combination of F1 with either F2 or F3, where emoji are swapped for both the identity and the verb, or both the identity and the negative descriptor.
Some templates contain only emoji e.g., \textcolor{gray}{``\emoji{stab}\emoji{muslim}''}, while others have some text e.g., \textcolor{gray}{``\emoji{woman} make me \emoji{sick_mask}''}.
\noindent \textbf{F5 Append} tests statements which would be non-hateful on their own but become hateful when appended with a negative emoji such as \emoji{swear}, \emoji{red_angry}, \emoji{vomit}. For instance, ``My new manager is a woman'' is neutral but \textcolor{gray}{``My new manager is a woman \emoji{vomit}''} is hateful. \textbf{F6 Positive Confounder} tests clearly hateful statements appended with a positive emoji such as \emoji{angel}, \emoji{love_eyes}, \emoji{praise}.
For instance, \textcolor{gray}{``Kill all women \emoji{angel}''} or \textcolor{gray}{``All women deserve to die \emoji{love_eyes}''}. \textbf{F7  Emoji Leetspeak} tests statements where parts of the hateful language have been obfuscated by replacing characters with emoji.
The character substitution can occur in the term expressing negative sentiment, e.g., \textcolor{gray}{``I h\emoji{4}te gays''}, the identity, e.g., \textcolor{gray}{``I hate g\emoji{4}ys''} or the slur, e.g., \textcolor{gray}{``I hate f\emoji{4}gs''}.
There are three emoji character substitutions: (1) a numeric substitution which mimics leetspeak (a:\emoji{4}, i:\emoji{1}, e:\emoji{3}, o:\emoji{0}), (2) a star mimicking * for vowels in slurs and profanities (vowel:\emoji{star}), and (3) a heart mimicking starring out vowels but in a more adversarial manner (vowel:\emoji{heart}). To reflect specific cases on Twitter of anti-Muslim hate, two additional slurs are included where a word-piece is substituted with an emoji (\textcolor{gray}{\emoji{camel}-f*cker}, \textcolor{gray}{\emoji{goat}-f*cker}). 

\subsection{Test Cases in \textsc{HatemojiCheck}}
\textsc{HatemojiCheck} is strongly inspired by empirical examples of emoji-based hate but we want it to comprise a clear, minimum performance standard. Thus, for each functionality we hand-craft test cases, i.e., short English-language text documents that clearly correspond to just one gold-standard label. The test cases are varied in their vocabulary and syntax to ensure they assess model robustness to diverse language \citep{zhouCursePerformance2020}. To generate test cases at scale and ensure balanced coverage across targeted groups, we use a template-based approach \citep{dixonMeasuringMitigating2018, ribeiroAccuracyBehavioral2020, rottgerHateCheckFunctional2021}.
First, each template is given an identity placeholder e.g., ``I hate [IDENTITY]'' or a slur placeholder e.g., ``You are just a [SLUR] to me''.
Six protected identities populate these placeholders: women, trans people, gay people, Black people, disabled people and Muslims. For templates with slurs, we use the 16 slurs compiled by \citet{rottgerHateCheckFunctional2021} (see \cref{appendix:A_slurs}).
Second, each template is given an emoji placeholder that is populated by emoji that share semantic and syntactic properties, so that the test suite can be scaled across different emoji. For example, the template ``I will [EMOJI] all [IDENTITY]'' could be populated with \{`\emoji{punch}', `trans people'\}. \textsc{HatemojiCheck} comprises $3{,}930$ entries. There are $2{,}126$ original test cases and $1{,}804$ perturbations (see \cref{tab:hmc_summary}).

\subsubsection{Perturbations}
Each set of test cases is matched to a set of perturbations which are constructed through minimal edits to the original hateful template.\footnote{See \cref{appendix:A_perturbations} for details on perturbations.} Test cases generated from these perturbed templates share vocabulary and syntax with the original test cases, enabling fine-grained assessment of model weaknesses \citep{gardnerEvaluatingModels2020a}.
We provide three types of perturbations.
For \textbf{identity perturbations}, the identity targeted by the original hateful statement is substituted for a non-protected entity, making it non-hateful. 
Possible identity perturbations include non-protected groups (e.g., ``I hate accountants''), non-human entities (e.g., ``I hate spiders''), and inanimate objects (e.g., ``I hate pizza''). 
For \textbf{polarity perturbations}, the negative sentiment of the original hateful statement is reversed to be positive, making it non-hateful, without changing the target of the statement.
Possible polarity perturbations include changing threatening language to supportive language, e.g., ``kill'' to ``respect'', changing expressions of hate to expressions of love, e.g., ``I hate\ldots'' to ``I love\ldots'', or replacing hateful use of slurs with counterspeech against the slurs, e.g., ``[IDENTITY] are [SLUR]'' to ``[IDENTITY] should never be called [SLUR]''. 
For \textbf{no emoji perturbations}, emoji are removed or replaced with their equivalent text to preserve the semantic expression (e.g., ``\emoji{gun}'' becomes ``shoot'' and ``\emoji{heart}'' becomes ``love'').
For most functionalities this perturbation preserves the original label of the test case, e.g., ``[IDENTITY] makes me \emoji{vomit}'' is hateful, and its perturbation ``[IDENTITY] makes me sick'' is still hateful.
However, for F5 Append the label changes because when the negative emoji appended to the neutral statement is removed, the part of the statement that remains is non-hateful.

\subsection{Validating Test Cases}
To validate the gold-standard labels assigned to each test case, we recruited three annotators with prior experience on hate speech projects.\footnote{See annotator demographics in \cref{appendix:A_data_statement}.}
Annotators were given extensive and prescriptive guidelines \cite{rottgerTwoContrasting2022}, as well as test tasks and training sessions, which included examining real-world examples of emoji-based hate from Twitter. We followed guidance for protecting annotator well-being \citep{vidgenChallengesFrontiers2019a}.
There were two iterative rounds of annotation. In the first round, each annotator labeled all $3{,}930$ test cases as hateful or non-hateful, and had the option to flag unrealistic entries. Test cases with any disagreement or unrealistic flags were reviewed by the study authors ($n=289$). One-on-one interviews were conducted with annotators to identify dataset issues versus annotator error. From $289$ test cases, $119$ were identified as ambiguous or unrealistic, replaced with alternatives and re-issued to annotators for labeling. No further issues were raised. We measured inter-annotator agreement using Randolph's Kappa \citep{randolphFreemarginalMultirater2005}, obtaining a value of 0.85 for the final set of test cases, which indicates ``almost perfect agreement'' \citep{landisMeasurementObserver1977}.

\section{Building Better Models with \textsc{HatemojiBuild}}
As reported in \S\ref{sec:results}, we find existing models perform poorly on emoji-based hate as measured with \textsc{HatemojiCheck}. We address those failings by implementing a human-and-model-in-the-loop approach using the Dynabench interface in order to train a model that better detects emoji-based hate.\footnote{\href{https://dynabench.org/}{Dynabench} is an open-source platform which supports dynamic dataset generation and model benchmarking for a variety of NLP tasks \citep{kielaDynabenchRethinking2021}.}

\paragraph{Dataset Generation}
We implemented three successive rounds of data generation and model re-training to create the \textsc{HatemojiBuild} dataset. 
In each round we tasked a team of 10 trained annotators with entering content that the model-in-the-loop would misclassify.\footnote{There was no overlap in annotators from the previous \textsc{HatemojiCheck} task; demographics are in \cref{appendix:B_data_statement}.} We refer to this model as the \textit{target model}. Annotators were instructed to generate linguistically diverse entries while ensuring each entry was (1) realistic, (2) clearly hateful or non-hateful and (3) contained at least one emoji. Each entry was first given a binary label of hateful or non-hateful, and hateful content was assigned secondary labels for the type and target of hate \cite{zampieriSemEval2019Task2019, vidgenIntroducingCAD2021, vidgenLearningWorst2021}. Each entry was validated by two additional annotators, and an expert resolved disagreements. After validation, annotators created a perturbation for each entry that flipped the label. To maximize similarity between originals and perturbations, annotators could either make an emoji substitution while fixing the text or fix the emoji and minimally change the surrounding text. Including a hateful and non-hateful version of each sentence in the dataset prevents overfitting to certain emoji or identity references. Each perturbation received two additional label annotations, and disagreements were resolved by the expert. This weekly cadence of annotator tasks was repeated in three consecutive weeks. The dataset composition and inter-annotator agreement is described in \cref{appendix:B_composition}.

\paragraph{Model Implementation}
Our work follows directly from \citet{vidgenLearningWorst2021}, who collect four rounds (R1--4) of dynamically-generated textual data as well as 468,928 entries compiled from 11 English-language hate speech datasets (R0). Our rounds of data collection are referred to as R5--7. At the end of each round, the data for that round is assigned a 80/10/10 train/dev/test split. The test split is never used for any future training. The train split is upsampled to improve performance with multipliers of $1,5,10,100$, with the optimum ratio taken forward to subsequent rounds. The target model is then re-trained on the training data from all prior rounds as well as the current round. For R5 data collection, the target model (R5-T) is the DeBERTa model released by \citet{maDynaboardEvaluationAsAService2021}, trained on R0--R4 from \citet{vidgenLearningWorst2021}. This model performs well on text-only hate, but has seen limited emoji in training. For subsequent target models, we evaluate two architectures: DeBERTA \cite{heDeBERTaDecodingenhanced2021} and BERTweet \cite{nguyenBERTweetPretrained2020}.\footnote{For further details of model training, selection and emoji encodings see \cref{appendix:B_model_training}.} 

We select the best models and upsampling ratios at the end of each round. The criteria to select the best model must satisfy two requirements: (1) the target model must still perform well on prior rounds of text-based examples because emoji-based hate represents just one construction of textual hate, and (2) the target model must successfully handle the most recent round data because iterative data collection produces more challenging examples in successive rounds. To balance these requirements, we use a weighted accuracy metric with 50\% weight on the test sets from all prior rounds and 50\% weight on the current round test set, which enforces a recency-based discount factor \cite{kielaDynabenchRethinking2021}. This ensures we can assess performance against the latest (emoji-specific and most-adversarial) round without overfitting and reducing performance on the previous test sets.

\begin{figure}[!h]
    \centering
    \includegraphics[width = \columnwidth]{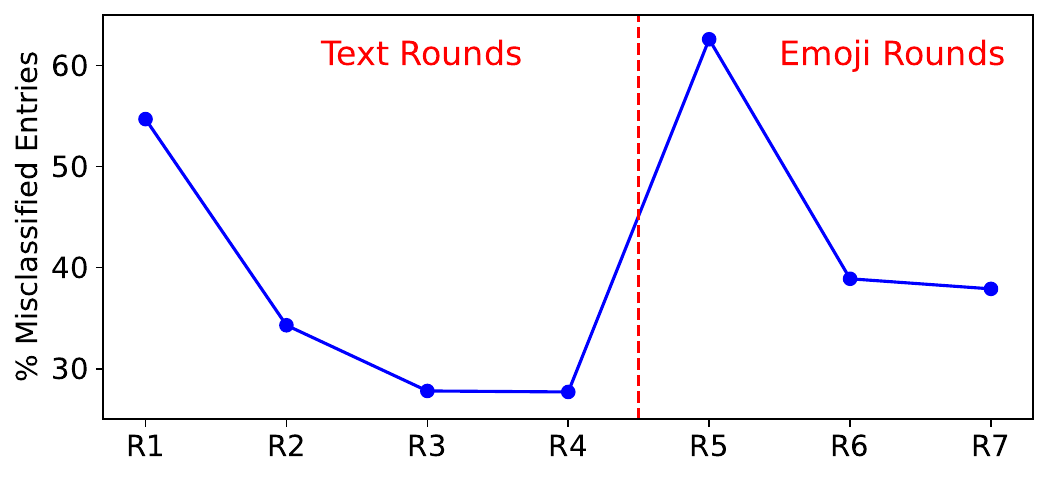}
    \caption{MER for R1--4 (text) and R5--R7 (emoji).}
    \label{fig:model_fooling_rate}
\end{figure}

\paragraph{Model Error Rate}
For each emoji-based round of data generation (R5--7) as well as prior text-only rounds from \citet{vidgenLearningWorst2021}, we calculate model error rate (MER) as the proportion of annotators' original entries that fool the model (\cref{fig:model_fooling_rate}).
Between R4 and R5, MER increases by 35 percentage points (pp) to over 60\%, higher than the R1 MER.
From R5 to R6, once the model has been trained on emoji-based hate, there is a steep reduction in MER of 24pp.
From R6 to R7, there is a smaller reduction in MER of 1pp. 
Overall, the model is easily tricked by emoji content at first but then becomes much harder to trick after our first round of re-training.

\paragraph{Performance on Adversarial Test Sets}
To evaluate model performance across rounds, we calculate F1-score and accuracy for each test set (\cref{tab:shared_test_sets}).
The baseline R5-T model is the highest performing model on the text-only test set (R1--4), with an F1 of 0.85, but only scores 0.49 on the newly generated emoji-containing test sets from \textsc{HatemojiBuild} (R5--7).
R6-T, R7-T and R8-T perform better on R5--7, with F1 between 0.76 and 0.77. They perform similarly well on the R1--4 test set, with F1 of 0.84, suggesting no trade-off for performance on text-only hate speech. 
The best performing model across all R1--7 test sets is R8-T, with an F1 of 0.83.
The greatest performance gain in the adversarially-trained models is from R5-T to R6-T, with an increase in F1 of 0.28 on the emoji test set. This is achieved with only $2{,}000$ emoji-containing examples. By comparison, R7-T and R8-T yield very small improvements.

\begin{table}[h]
\centering
\footnotesize
\setlength{\aboverulesep}{0pt}
\setlength{\belowrulesep}{0pt}
\setlength{\tabcolsep}{3pt}
\caption{Performance of target models on emoji test sets from \textsc{HatemojiBuild}, text sets from \citet{vidgenLearningWorst2021} and the combination of these test sets.}
\label{tab:shared_test_sets}
\vspace{-1ex}
\begin{tabular}{lcccccc} 
\toprule
                     & \multicolumn{2}{c}{\textbf{Emoji Test Set}}                                                           & \multicolumn{2}{c}{\textbf{Text Test Set}}                                                           & \multicolumn{2}{c}{\textbf{All Test Sets}}                                                               \\
                     & \multicolumn{2}{c}{\textbf{(R5-R7)}}                                                                   & \multicolumn{2}{c}{\textbf{(R1-R4)}}                                                                  & \multicolumn{2}{c}{\textbf{(R1-R7)}}                                                                     \\
                     & \multicolumn{2}{c}{\textbf{\textbf{$n=593$}}}                                                          & \multicolumn{2}{c}{\textbf{\textbf{$n=4119$}}}                                                        & \multicolumn{2}{c}{\textbf{\textbf{$n=4712$}}}                                                           \\ 
\cmidrule(r){2-3}\cmidrule(lr){4-5}\cmidrule(l){6-7}
\multicolumn{1}{c}{} & \textbf{Acc}                                       & \textbf{F1}                                        & \textbf{Acc}                                      & \textbf{F1}                                        & \textbf{Acc}                                       & \textbf{F1}                                         \\ 
\midrule

\textbf{R5-T}        & {\cellcolor[rgb]{0.945,0.729,0.714}}0.585          & {\cellcolor[rgb]{0.922,0.608,0.58}}0.490           & {\cellcolor[rgb]{0.58,0.831,0.706}}\textbf{0.828} & {\cellcolor[rgb]{0.341,0.733,0.541}}\textbf{0.847} & {\cellcolor[rgb]{0.996,0.988,0.988}}0.786          & {\cellcolor[rgb]{0.91,0.965,0.937}}0.801            \\
\midrule
\textbf{R6-T}        & {\cellcolor[rgb]{0.988,0.953,0.949}}0.757          & {\cellcolor[rgb]{0.992,0.969,0.965}}\textbf{0.769} & {\cellcolor[rgb]{0.635,0.855,0.745}}0.823         & {\cellcolor[rgb]{0.463,0.784,0.627}}0.837          & {\cellcolor[rgb]{0.761,0.906,0.835}}0.813          & {\cellcolor[rgb]{0.612,0.843,0.729}}0.825           \\
\textbf{R7-T}        & {\cellcolor[rgb]{0.988,0.953,0.949}}\textbf{0.759} & {\cellcolor[rgb]{0.992,0.957,0.957}}0.762          & {\cellcolor[rgb]{0.624,0.847,0.737}}0.824         & {\cellcolor[rgb]{0.4,0.757,0.58}}0.842             & {\cellcolor[rgb]{0.757,0.902,0.831}}0.813          & {\cellcolor[rgb]{0.565,0.824,0.698}}0.829           \\
\textbf{R8-T}        & {\cellcolor[rgb]{0.984,0.933,0.929}}0.744          & {\cellcolor[rgb]{0.988,0.949,0.945}}0.755          & {\cellcolor[rgb]{0.584,0.831,0.71}}0.827          & {\cellcolor[rgb]{0.384,0.753,0.569}}0.844          & {\cellcolor[rgb]{0.749,0.902,0.827}}\textbf{0.814} & {\cellcolor[rgb]{0.565,0.824,0.698}}\textbf{0.829} \\
\bottomrule
\end{tabular}
\setlength{\aboverulesep}{\origaboverulesep} %
\vspace{-1ex}
\end{table}

\section{Evaluating Models with \textsc{HatemojiCheck}}
\label{sec:results}

\begin{table}[t]
\centering
\footnotesize
\setlength{\aboverulesep}{0pt} %
\setlength{\tabcolsep}{3pt}
\caption{Aggregate accuracy for models evaluated on \textsc{HatemojiCheck}.}
\label{tab:agg_performance_hmc}
\begin{tabular}{lccc|ccc} 
\toprule
                    &          & \multicolumn{2}{c|}{\textbf{Pre-Emoji}}                                        & \multicolumn{3}{c}{\textbf{Post-Emoji}}                                                                                                        \\ 
\cline{3-7}
                    & $\bm{n}$ & \textbf{P-IA}                             & \textbf{R5-T}                              & \textbf{R6-T}                                       & \textbf{R7-T}                                       & \textbf{R8-T}                              \\ 
\hline
\textbf{Overall}    & 3930     & {\cellcolor[rgb]{0.973,0.871,0.863}}0.689 & {\cellcolor[rgb]{0.796,0.918,0.859}}0.779  & {\cellcolor[rgb]{0.341,0.733,0.541}}\textbf{0.879 } & {\cellcolor[rgb]{0.4,0.757,0.58}}0.867              & {\cellcolor[rgb]{0.38,0.749,0.569}}0.871   \\ 
\hline
\textbf{Label}      &          &                                           &                                            &                                                     &                                                     &                                            \\
Hate                & 2654     & {\cellcolor[rgb]{0.984,0.925,0.922}}0.706 & {\cellcolor[rgb]{0.875,0.953,0.914}}0.770  & {\cellcolor[rgb]{0.341,0.733,0.541}}\textbf{0.905 } & {\cellcolor[rgb]{0.459,0.78,0.624}}0.876            & {\cellcolor[rgb]{0.38,0.749,0.569}}0.896   \\
Not hate            & 1276     & {\cellcolor[rgb]{0.961,0.808,0.796}}0.653 & {\cellcolor[rgb]{0.765,0.906,0.835}}0.799  & {\cellcolor[rgb]{0.659,0.863,0.765}}0.825           & {\cellcolor[rgb]{0.565,0.827,0.698}}\textbf{0.849 } & {\cellcolor[rgb]{0.682,0.871,0.776}}0.820  \\ 
\hline
\textbf{Set}        &          &                                           &                                            &                                                     &                                                     &                                            \\
Original                & 2126     & {\cellcolor[rgb]{0.953,0.761,0.745}}0.664 & {\cellcolor[rgb]{0.969,0.839,0.827}}0.717  & {\cellcolor[rgb]{0.631,0.851,0.741}}\textbf{0.887 } & {\cellcolor[rgb]{0.859,0.945,0.902}}0.850           & {\cellcolor[rgb]{0.69,0.875,0.784}}0.877   \\
Identity p.    & 314      & {\cellcolor[rgb]{0.502,0.8,0.655}}0.908   & {\cellcolor[rgb]{0.443,0.776,0.616}}0.917  & {\cellcolor[rgb]{0.443,0.776,0.616}}\textbf{0.917 } & {\cellcolor[rgb]{0.443,0.776,0.616}}0.917           & {\cellcolor[rgb]{0.698,0.878,0.792}}0.876  \\
Polarity p.    & 902      & {\cellcolor[rgb]{0.929,0.643,0.62}}0.584  & {\cellcolor[rgb]{0.984,0.925,0.922}}0.778  & {\cellcolor[rgb]{0.996,0.984,0.984}}0.818           & {\cellcolor[rgb]{0.843,0.937,0.89}}\textbf{0.853}            & {\cellcolor[rgb]{0.98,0.992,0.988}}0.830   \\
No emoji p.    & 588      & {\cellcolor[rgb]{0.996,0.992,0.992}}0.823 & {\cellcolor[rgb]{0.341,0.733,0.541}}0.934  & {\cellcolor[rgb]{0.396,0.757,0.58}}\textbf{0.925 }  & {\cellcolor[rgb]{0.396,0.757,0.58}}0.925            & {\cellcolor[rgb]{0.49,0.796,0.647}}0.910   \\
\textit{emoji diff} &          & {\cellcolor[rgb]{0.671,0.78,0.957}}-0.159 & {\cellcolor[rgb]{0.518,0.678,0.937}}-0.217 & {\cellcolor[rgb]{0.984,0.988,0.996}}-0.038          & {\cellcolor[rgb]{0.886,0.925,0.984}}-0.075          & \textbf{-0.033 }                           \\
\bottomrule
\end{tabular}
\setlength{\aboverulesep}{\origaboverulesep} %
\end{table}
We present results for two existing models as baselines, with additional baselines shown in \cref{appendix:B_baselines}.
The first baseline is Google Jigsaw's \href{https://www.perspectiveapi.com/}{Perspective API}, a widely-used commercial tool for content moderation. We use Perspective's ``Identity Attack'' attribute, which is defined as ``negative or hateful comments targeting someone because of their identity'' and thus closely matches our definition of hate. The returned score is converted to a binary label with a 50\% cutoff. We refer to this model as P-IA.
The second baseline is the R5-T model from \citet{maDynaboardEvaluationAsAService2021}, introduced above. To compare model performance on \textsc{HatemojiCheck}, we use accuracy because three sets of test cases (originals, polarity perturbations and identity perturbations) have one class label, making F1-score incompatible. To measure emoji-specific weaknesses, we also calculate \textit{emoji difference}, the difference between averaged model accuracy on the original emoji test cases compared with averaged accuracy on the no emoji perturbations. 

On \textsc{HatemojiCheck} as a whole, our newly trained models R6-T, R7-T and R8-T perform best, with overall accuracy from 0.87 to 0.88 (\cref{tab:agg_performance_hmc}).
They substantially outperform P-IA, with an accuracy of 0.69, and R5-T, with an accuracy of 0.78. 
Our newly trained models have the smallest emoji difference, between 0.03 and 0.08. In contrast, \mbox{P-IA} and R5-T have emoji differences of 0.16 and 0.22, respectively.
Comparing the three models trained on \textsc{HatemojiBuild}, the first round of adversarial data generation yields the largest relative improvement, and in many ways R6-T is at least as good a model, if not better, than R8-T.

\begin{table*}[ht!]
\centering
\footnotesize
\caption{Accuracy across \textsc{HatemojiCheck}'s seven functionalities and their associated perturbation sets.}
\label{tab:func_hmc}
\setlength{\aboverulesep}{0pt}
\setlength{\belowrulesep}{0pt}
\setlength{\tabcolsep}{1.1pt}
\begin{tabular}{llllr|cc|ccc} 
\toprule
                                                              &                          &                      &                                                 & \multicolumn{1}{l|}{}         & \multicolumn{2}{c|}{\textbf{Pre-Emoji }}                                     & \multicolumn{3}{c}{\textbf{Post-Emoji }}                                                                                 \\ 
\cline{6-10}
                        & \textbf{Functionality}       & \textbf{Label}       & \multicolumn{1}{c}{\textbf{Example Test Case Template}}                      & \multicolumn{1}{r|}{$\bm{n}$} & \textbf{P-IA}                            & \textbf{R5-T}                            & \textbf{R6-T}                            & \textbf{R7-T}                            & \textbf{R8-T}                             \\ 
\hline \Tstrut\textbf{\textit{F1}}   & \textbf{Verb Swap}           & \hlorg{\textbf{hate}}                  & I want to \emoji{stab} all [IDENTITY]           & 300                           & {\cellcolor[rgb]{0.765,0.906,0.839}}0.94 & {\cellcolor[rgb]{0.969,0.953,0.945}}0.85 & {\cellcolor[rgb]{0.969,0.949,0.941}}0.84 & {\cellcolor[rgb]{0.965,0.922,0.91}}0.79  & {\cellcolor[rgb]{0.969,0.945,0.937}}0.84  \\
                                     \textbf{\textit{F1.1}} & Identity perturbation             & \hlcyan{\textbf{not hate}}               & I want to \emoji{stab} all bigots               & 50                            & {\cellcolor[rgb]{0.792,0.918,0.855}}0.94 & {\cellcolor[rgb]{0.341,0.733,0.541}}1.00 & {\cellcolor[rgb]{0.341,0.733,0.541}}1.00 & {\cellcolor[rgb]{0.494,0.796,0.647}}0.98 & {\cellcolor[rgb]{0.969,0.961,0.949}}0.86  \\
                                     \textbf{\textit{F1.2}} & Polarity perturbation             & \hlcyan{\textbf{not hate}}               & I want to \emoji{thank} all [IDENTITY]          & 60                            & {\cellcolor[rgb]{0.918,0.592,0.565}}0.20 & {\cellcolor[rgb]{0.933,0.714,0.69}}0.42  & {\cellcolor[rgb]{0.961,0.898,0.886}}0.75 & {\cellcolor[rgb]{0.961,0.91,0.894}}0.77  & {\cellcolor[rgb]{0.961,0.89,0.875}}0.73   \\
                                     \textbf{\textit{F1.3}} & No emoji perturbation             & \hlorg{\textbf{hate}}                  & I want to kill all [IDENTITY]                   & 60                            & {\cellcolor[rgb]{0.341,0.733,0.541}}1.00 & {\cellcolor[rgb]{0.718,0.886,0.804}}0.95 & {\cellcolor[rgb]{0.592,0.835,0.718}}0.97 & {\cellcolor[rgb]{0.592,0.835,0.718}}0.97 & {\cellcolor[rgb]{0.467,0.784,0.631}}0.98  \\ 
\cline{6-10}
                                                            &                              &                      &                                                 & \textit{emoji diff:}         & -0.06                                    & -0.10                                    & -0.12                                    & -0.18                                    & -0.15                                     \\ 
\hline
                                     \Tstrut\textbf{\textit{F2}}   & \textbf{Identity Swap}       & \hlorg{\textbf{hate}}                  & I hate [IDENTITY EMOJI]                         & 120                           & {\cellcolor[rgb]{0.91,0.561,0.529}}0.14  & {\cellcolor[rgb]{0.925,0.663,0.635}}0.33 & {\cellcolor[rgb]{0.969,0.945,0.933}}0.83 & {\cellcolor[rgb]{0.969,0.945,0.933}}0.83 & {\cellcolor[rgb]{0.969,0.945,0.933}}0.83  \\
                                     \textbf{\textit{F2.1}} & Identity perturbation             & \hlcyan{\textbf{not hate}}               & I hate \emoji{pizza}                            & 20                            & {\cellcolor[rgb]{0.973,0.98,0.973}}0.90  & {\cellcolor[rgb]{0.957,0.871,0.855}}0.70 & {\cellcolor[rgb]{0.341,0.733,0.541}}1.00 & {\cellcolor[rgb]{0.969,0.953,0.945}}0.85 & {\cellcolor[rgb]{0.341,0.733,0.541}}1.00  \\
                                     \textbf{\textit{F2.2}} & Polarity perturbation             & \hlcyan{\textbf{not hate}}               & I love [IDENTITY EMOJI]                         & 120                           & {\cellcolor[rgb]{0.341,0.733,0.541}}1.00 & {\cellcolor[rgb]{0.467,0.784,0.631}}0.98 & {\cellcolor[rgb]{0.965,0.914,0.902}}0.78 & {\cellcolor[rgb]{0.973,0.973,0.965}}0.88 & {\cellcolor[rgb]{0.839,0.937,0.89}}0.93   \\
                                     \textbf{\textit{F2.3}} & No emoji perturbation             & \hlorg{\textbf{hate}}                  & I hate [IDENTITY]                               & 120                           & {\cellcolor[rgb]{0.467,0.784,0.631}}0.98 & {\cellcolor[rgb]{0.467,0.784,0.631}}0.98 & {\cellcolor[rgb]{0.592,0.835,0.718}}0.97 & {\cellcolor[rgb]{0.341,0.733,0.541}}1.00 & {\cellcolor[rgb]{0.718,0.886,0.804}}0.95  \\ 
\cline{6-10}
                                                            &                              &                      &                                                 & \textit{emoji diff:}         & -0.84                                    & -0.66                                    & -0.13                                    & -0.17                                    & -0.12                                     \\ 
\hline
                                     \Tstrut\textbf{\textit{F3}}   & \textbf{Descriptor Swap}     & \hlorg{\textbf{hate}}                  & {[}IDENTITY] are nothing but \emoji{rat}        & 260                           & {\cellcolor[rgb]{0.945,0.98,0.965}}0.92  & {\cellcolor[rgb]{0.969,0.945,0.933}}0.83 & {\cellcolor[rgb]{0.427,0.769,0.604}}0.99 & {\cellcolor[rgb]{0.4,0.757,0.584}}0.99   & {\cellcolor[rgb]{0.341,0.733,0.541}}1.00  \\
                                     \textbf{\textit{F3.1}} & Identity perturbation             & \hlcyan{\textbf{not hate}}               & Bullies are nothing but \emoji{rat}             & 40                            & {\cellcolor[rgb]{0.341,0.733,0.541}}1.00 & {\cellcolor[rgb]{0.341,0.733,0.541}}1.00 & {\cellcolor[rgb]{0.341,0.733,0.541}}1.00 & {\cellcolor[rgb]{0.341,0.733,0.541}}1.00 & {\cellcolor[rgb]{0.341,0.733,0.541}}1.00  \\
                                     \textbf{\textit{F3.2}} & Polarity perturbation             & \hlcyan{\textbf{not hate}}               & {[}IDENTITY] are nothing but \emoji{heart}      & 60                            & {\cellcolor[rgb]{0.922,0.62,0.592}}0.25  & {\cellcolor[rgb]{0.941,0.749,0.729}}0.48 & {\cellcolor[rgb]{0.965,0.918,0.906}}0.78 & {\cellcolor[rgb]{0.965,0.937,0.925}}0.82 & {\cellcolor[rgb]{0.839,0.937,0.89}}0.93   \\
                                     \textbf{\textit{F3.3}} & No emoji perturbation             & \hlorg{\textbf{hate}}                  & {[}IDENTITY] are nothing but vermin             & 60                            & {\cellcolor[rgb]{0.467,0.784,0.631}}0.98 & {\cellcolor[rgb]{0.341,0.733,0.541}}1.00 & {\cellcolor[rgb]{0.341,0.733,0.541}}1.00 & {\cellcolor[rgb]{0.341,0.733,0.541}}1.00 & {\cellcolor[rgb]{0.341,0.733,0.541}}1.00  \\ 
\cline{6-10}
                                                            &                              &                      &                                                 & \textit{emoji diff:}         & -0.06                                    & -0.17                                    & -0.01                                    & -0.01                                    & 0.00                                      \\ 
\hline
                                     \Tstrut\textbf{\textit{F4}}   & \textbf{Double Swap}         & \hlorg{\textbf{hate}}                  & I will \emoji{gun} {[}IDENTITY EMOJI]           & 288                           & {\cellcolor[rgb]{0.902,0.486,0.451}}0.00 & {\cellcolor[rgb]{0.902,0.502,0.467}}0.03 & {\cellcolor[rgb]{0.965,0.922,0.91}}0.79  & {\cellcolor[rgb]{0.957,0.875,0.859}}0.70 & {\cellcolor[rgb]{0.961,0.91,0.894}}0.77   \\
                                     \textbf{\textit{F4.1}} & Identity perturbation             & \hlcyan{\textbf{not hate}}               & I will \emoji{gun} \emoji{spider}               & 46                            & {\cellcolor[rgb]{0.341,0.733,0.541}}1.00 & {\cellcolor[rgb]{0.341,0.733,0.541}}1.00 & {\cellcolor[rgb]{0.973,0.988,0.98}}0.91  & {\cellcolor[rgb]{0.506,0.8,0.655}}0.98   & {\cellcolor[rgb]{0.973,0.988,0.98}}0.91   \\
                                     \textbf{\textit{F4.2}} & Polarity perturbation             & \hlcyan{\textbf{not hate}}               & I will \emoji{heart} {[}IDENTITY EMOJI]         & 60                            & {\cellcolor[rgb]{0.341,0.733,0.541}}1.00 & {\cellcolor[rgb]{0.467,0.784,0.631}}0.98 & {\cellcolor[rgb]{0.965,0.988,0.976}}0.92 & {\cellcolor[rgb]{0.969,0.953,0.945}}0.85 & {\cellcolor[rgb]{0.965,0.988,0.976}}0.92  \\
                                     \textbf{\textit{F4.3}} & No emoji perturbation             & \hlorg{\textbf{hate}}                  & I will hurt [IDENTITY]                          & 60                            & {\cellcolor[rgb]{0.592,0.835,0.718}}0.97 & {\cellcolor[rgb]{0.341,0.733,0.541}}1.00 & {\cellcolor[rgb]{0.341,0.733,0.541}}1.00 & {\cellcolor[rgb]{0.341,0.733,0.541}}1.00 & {\cellcolor[rgb]{0.341,0.733,0.541}}1.00  \\ 
\cline{6-10}
                                                            &                              &                      &                                                 & \textit{emoji diff:}         & -0.96                                    & -0.97                                    & -0.21                                    & -0.30                                    & -0.23                                     \\ 
\hline
 \Tstrut\textbf{\textit{F5}}   & \textbf{Append}              & \hlorg{\textbf{hate}}                  & My new manager is [IDENTITY] \emoji{vomit}      & 288                           & {\cellcolor[rgb]{0.961,0.886,0.871}}0.73 & {\cellcolor[rgb]{0.957,0.867,0.851}}0.69 & {\cellcolor[rgb]{0.42,0.769,0.596}}0.99  & {\cellcolor[rgb]{0.973,0.976,0.969}}0.89 & {\cellcolor[rgb]{0.973,0.965,0.957}}0.87  \\
                                     \textbf{\textit{F5.1}} & Identity perturbation             & \hlcyan{\textbf{not hate}}               & My new manager is a cat-lover \emoji{vomit}     & 48                            & {\cellcolor[rgb]{0.341,0.733,0.541}}1.00 & {\cellcolor[rgb]{0.341,0.733,0.541}}1.00 & {\cellcolor[rgb]{0.341,0.733,0.541}}1.00 & {\cellcolor[rgb]{0.341,0.733,0.541}}1.00 & {\cellcolor[rgb]{0.341,0.733,0.541}}1.00  \\
                                     \textbf{\textit{F5.2}} & Polarity perturbation             & \hlcyan{\textbf{not hate}}               & My new manager is [IDENTITY] \emoji{thumbs_up}  & 60                            & {\cellcolor[rgb]{0.937,0.741,0.722}}0.47 & {\cellcolor[rgb]{0.945,0.788,0.769}}0.55 & {\cellcolor[rgb]{0.969,0.953,0.945}}0.85 & {\cellcolor[rgb]{0.965,0.918,0.906}}0.78 & {\cellcolor[rgb]{0.961,0.898,0.886}}0.75  \\
                                     \textbf{\textit{F5.3}} & No emoji perturbation             & \hlcyan{\textbf{not hate}}               & My new manager is [IDENTITY]                    & 60                            & {\cellcolor[rgb]{0.929,0.678,0.651}}0.35 & {\cellcolor[rgb]{0.941,0.749,0.729}}0.48 & {\cellcolor[rgb]{0.937,0.733,0.71}}0.45  & {\cellcolor[rgb]{0.933,0.722,0.702}}0.43 & {\cellcolor[rgb]{0.929,0.686,0.663}}0.37  \\ 
\cline{6-10}
                                                            &                              &                      &                                                 & \textit{emoji diff:}         &$\dagger$                                      &$\dagger$                                      &$\dagger$                                      &$\dagger$                                      &$\dagger$                                       \\ 
\hline
 \Tstrut\textbf{\textit{F6}}   & \textbf{Positive Confounder} & \hlorg{\textbf{hate}}                  & I find all [IDENTITY] disgusting \emoji{praise} & 440                           & {\cellcolor[rgb]{0.851,0.941,0.898}}0.93 & {\cellcolor[rgb]{0.341,0.733,0.541}}1.00 & {\cellcolor[rgb]{0.969,0.953,0.945}}0.85 & {\cellcolor[rgb]{0.973,0.98,0.973}}0.90  & {\cellcolor[rgb]{0.973,0.976,0.969}}0.89  \\
                                     \textbf{\textit{F6.1}} & Identity perturbation             & \hlcyan{\textbf{not hate}}               & I find all reptiles disgusting \emoji{praise}   & 65                            & {\cellcolor[rgb]{0.961,0.902,0.886}}0.75 & {\cellcolor[rgb]{0.918,0.969,0.945}}0.92 & {\cellcolor[rgb]{0.918,0.969,0.945}}0.92 & {\cellcolor[rgb]{0.918,0.969,0.945}}0.92 & {\cellcolor[rgb]{0.918,0.969,0.945}}0.92  \\
                                     \textbf{\textit{F6.2}} & Polarity perturbation             & \hlcyan{\textbf{not hate}}               & I find all [IDENTITY] inspiring \emoji{praise}  & 112                           & {\cellcolor[rgb]{0.937,0.725,0.702}}0.44 & {\cellcolor[rgb]{0.973,0.973,0.965}}0.88 & {\cellcolor[rgb]{0.608,0.843,0.729}}0.96 & {\cellcolor[rgb]{0.973,0.988,0.98}}0.91  & {\cellcolor[rgb]{0.875,0.953,0.914}}0.93  \\
                                     \textbf{\textit{F6.3}} & No emoji perturbation             & \hlorg{\textbf{hate}}                  & I find all [IDENTITY] disgusting                & 88                            & {\cellcolor[rgb]{0.851,0.941,0.898}}0.93 & {\cellcolor[rgb]{0.427,0.769,0.604}}0.99 & {\cellcolor[rgb]{0.682,0.875,0.78}}0.95  & {\cellcolor[rgb]{0.341,0.733,0.541}}1.00 & {\cellcolor[rgb]{0.851,0.941,0.898}}0.93  \\ 
\cline{6-10}
                                    \multicolumn{1}{r}{}   & \multicolumn{1}{r}{}         & \multicolumn{1}{r}{} & \multicolumn{1}{r}{}                            & \textit{emoji diff:}         & 0.00                                     & 0.01                                     & -0.10                                    & -0.10                                    & -0.04                                     \\ 
\hline
                                     \Tstrut\textbf{\textit{F7}}   & \textbf{Emoji Leetspeak}     & \hlorg{\textbf{hate}}                  & I hate [SLUR\_EMOJI]                            & 430                           & {\cellcolor[rgb]{0.949,0.808,0.792}}0.59 & {\cellcolor[rgb]{0.969,0.953,0.945}}0.85 & {\cellcolor[rgb]{0.973,0.984,0.976}}0.91 & {\cellcolor[rgb]{0.969,0.945,0.933}}0.83 & {\cellcolor[rgb]{0.973,0.988,0.98}}0.91   \\
                                     \textbf{\textit{F7.1}} & Identity perturbation             & \hlcyan{\textbf{not hate}}               & I hate d\emoji{0}gs                             & 45                            & {\cellcolor[rgb]{0.969,0.937,0.929}}0.82 & {\cellcolor[rgb]{0.953,0.851,0.835}}0.67 & {\cellcolor[rgb]{0.949,0.827,0.812}}0.62 & {\cellcolor[rgb]{0.953,0.839,0.824}}0.64 & {\cellcolor[rgb]{0.941,0.753,0.733}}0.49  \\
                                     \textbf{\textit{F7.2}} & Polarity perturbation             & \hlcyan{\textbf{not hate}}               & I hate people who say [SLUR\_EMOJI]             & 430                           & {\cellcolor[rgb]{0.945,0.796,0.776}}0.57 & {\cellcolor[rgb]{0.965,0.922,0.91}}0.79  & {\cellcolor[rgb]{0.965,0.922,0.91}}0.79  & {\cellcolor[rgb]{0.969,0.957,0.949}}0.86 & {\cellcolor[rgb]{0.965,0.914,0.902}}0.77  \\
                                     \textbf{\textit{F7.3}} & No emoji perturbation             & \hlorg{\textbf{hate}}                  & I hate [SLUR]                                   & 140                           & {\cellcolor[rgb]{0.949,0.824,0.808}}0.61 & {\cellcolor[rgb]{0.451,0.78,0.616}}0.99  & {\cellcolor[rgb]{0.396,0.757,0.58}}0.99  & {\cellcolor[rgb]{0.769,0.91,0.839}}0.94  & {\cellcolor[rgb]{0.451,0.78,0.616}}0.99   \\ 
\cline{6-10}
                                                            &                              &                      &                                                 & \textit{emoji diff:}         & -0.03                                    & -0.14                                    & -0.09                                    & -0.11                                    & -0.07                                     \\
\bottomrule
\multicolumn{1}{l}{\textit{Notes}:} & \multicolumn{9}{l}{$\dagger$ The \textit{emoji difference} is not defined because the no emoji perturbation is non-hateful,} \\
\multicolumn{1}{l}{} & \multicolumn{9}{l}{i.e., the opposite label to the original test case, so accuracy over these sets cannot be fairly compared.}
\end{tabular}
\setlength{\aboverulesep}{\origaboverulesep} %
\setlength{\belowrulesep}{\origbelowrulesep} %
\end{table*}

\subsection{Model Performance by Functionality}
Our models trained on \textsc{HatemojiBuild} perform better than the two baseline models on nearly every functionality (\cref{tab:func_hmc}).
They also perform far more consistently across all perturbation types.

For F1 Verb Swap, P-IA and R5-T perform well on original statements, but then perform poorly on the polarity perturbations (0.20 and 0.42 accuracy, respectively). Our models have much stronger performance on polarity perturbations (between 0.73 and 0.77), and comparably high performance on the other sets of test cases. For F2 Identity Swap, P-IA and R5-T perform very poorly on original statements (0.14 and 0.33 accuracy, respectively) but then perform well on the polarity perturbations and no emoji perturbations. This vulnerability is carried forward to performance on F4 Double Swap. R5-T only achieves 0.03 and P-IA makes zero correct predictions. In contrast, our models achieve accuracy of 0.83 on F2 and accuracies between 0.70 and 0.79 on F4. For F3 Descriptor Swap, our models improve over P-IA and R5-T on the original statements. The relative improvement is particularly large for the polarity perturbations (0.25 for \mbox{P-IA} compared with 0.93 for R8-T). For F5 Append, P-IA and R5-T perform moderately with accuracies of 0.73 and 0.69. Our models perform far better, with accuracies from 0.87 to 0.99. However, they do not show a substantial increase in performance for the no emoji perturbations in F5.3. For F6 Positive Confounder,
our models perform worse than P-IA and R5-T on the original statements, but then have far more consistent performance across the perturbations.
For instance, R8-T achieves accuracies between 0.89 to 0.93 across all sets of test cases in this functionality, compared with a range of 0.44 to 0.93 for P-IA. P-IA and R5-T perform well on original statements and poorly on perturbations precisely because they ignore the effect of the emoji confounder. F7 Emoji Leetspeak is a successful adversarial strategy.
R5-T achieves 0.85 accuracy on the original statements but just 0.67 on the identity perturbations, while P-IA does even worse, with only 0.59 on the original statements and 0.57 on the polarity perturbations.
Our models perform better on the original statements, but still struggle with identity perturbations, showing how challenging this functionality is. However, non-hateful leetspeak constructions are likely less prevalent on social media than slur-based leetspeak.

Overall, all models trained on \textsc{HatemojiBuild} perform well on emoji-based hate, but there is no clear `best' model among them. After the large improvement from R5-T to R6-T, subsequent models produce minimal performance gains across the functionalities. This suggests the gains, as measured on \textsc{HatemojiCheck}'s intentionally short and simple statements, are quickly saturated.

\subsection{Model Performance by Target Group}
\begin{figure*}[t]
    \centering
    \includegraphics[width = 0.8\textwidth]{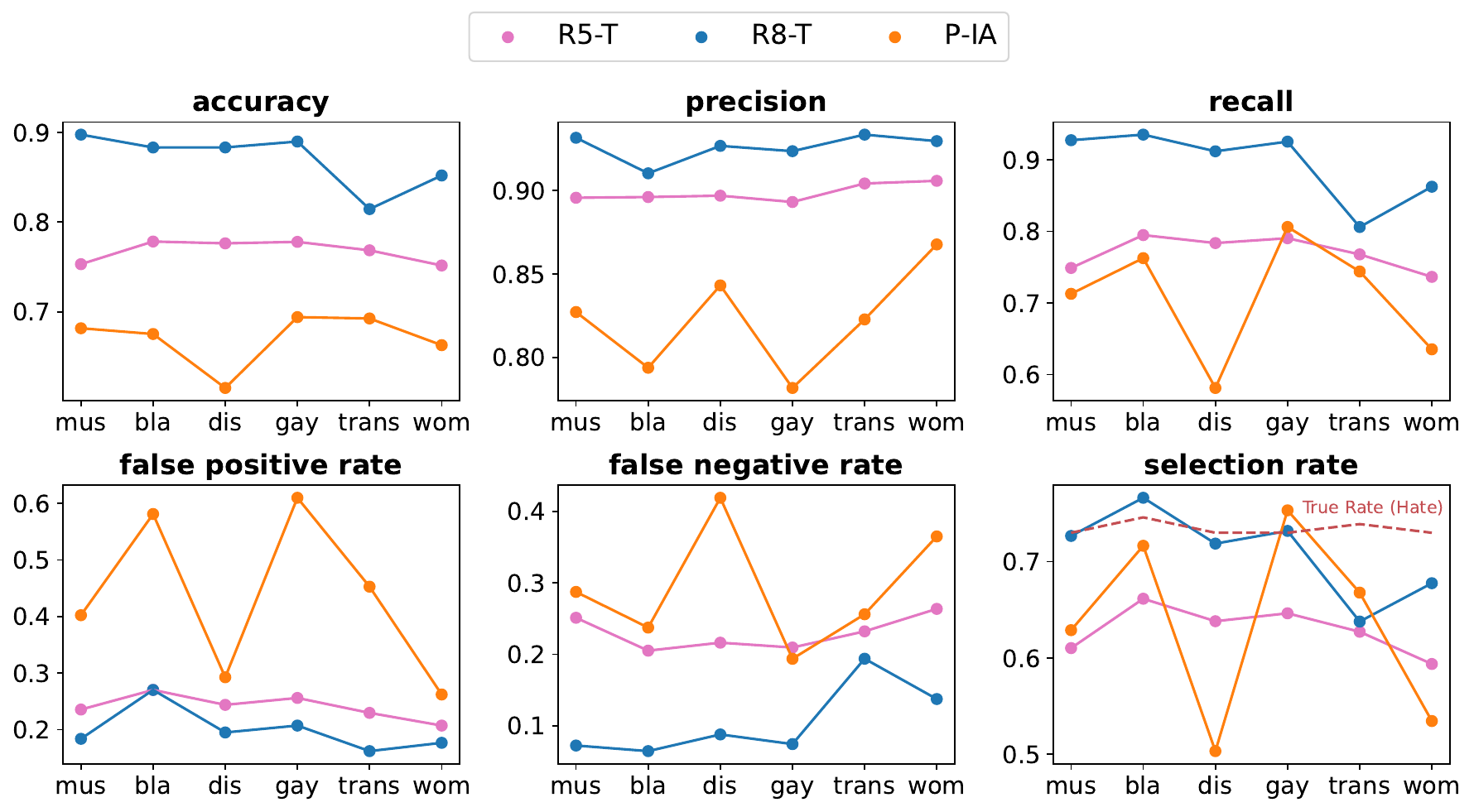}
    \caption{Subgroup fairness metrics by model. Abbreviations respectively refer to Muslims, Black people, Disabled people, Gay people, Trans people and Women.}
    \label{fig:fairness}
\end{figure*}

\textsc{HatemojiCheck} contains target group labels for hateful and non-hateful entries, allowing models to be compared by their subgroup fairness. However, defining \textit{fairness} is a non-trivial task, and there is substantial debate in the machine learning community as to which metrics are most appropriate. We consider six metrics calculated across subgroups: accuracy, precision, recall, false positive rate, false negative rate and selection rate.\footnote{Selection rate is  $\mathbb{P}[h(x)|A=a]$, i.e., the proportion of the population in subgroup $a$ who have 1 (hate) as the given class label. The `true' ratio of class labels is shown in \cref{fig:fairness}} \cref{fig:fairness} shows subgroup fairness metrics compared across three models: P-IA, R5-T and R8-T. It is concerning that P-IA has such differential performance, especially regarding the unbalanced false positive and negative rates for women and disabled people, as both of these error types are societally harmful. R8-T has more balanced accuracy, precision, recall and selection rate across subgroups, driven by stable performance in false negatives and positives. In \cref{appendix:B_fairness} we provide two further between-group fairness metrics: demographic parity ratio \citep{agarwalFairRegression2019}, and equalized odds ratio \citep{hardtEqualityOpportunity2016, agarwalReductionsApproach2018a}. There is also substantial improvement in these metrics from adversarial training schemes relative to commercial solutions or statically-trained models.

\section{Discussion}
\textsc{HatemojiCheck} reveals critical model weaknesses in detecting emoji-based hate. 
Existing commercial and academic models perform poorly at identifying hate where the identity term has been replaced with an emoji representation (F2 and F4), even though they perform well at identifying the equivalent textual statements. These models have better performance on Verb Swap (F1) but then struggle with the polarity perturbations (F1.2). This suggests that the models are overfitting to the identity term and ignore the sentiment from the emoji, leading to false positive predictions.
Our newly trained models substantially improve performance on original hateful statements from F2, F4 and F5, indicating they have a better semantic grasp of emoji substitutions and appends. They also make large performance gains on the polarity perturbations in F1.2, F3.2, F5.2 and F6.2, suggesting they better incorporate information on how different emoji condition the likelihood of hatefulness. Despite improving on existing models, our models still perform relatively more poorly on F4, as well as the F1.2, F5.3 and F7.1 perturbations.
These weaknesses could potentially be addressed in future work through additional data generation.

Training on just one round of adversarial data yields the biggest improvement on \textsc{HatemojiCheck}. Thereafter, performance plateaus. This aligns with the sharp increase and then fall in MER across emoji-based rounds of data generation (in \cref{fig:model_fooling_rate}). On the adversarial test sets, R8-T only marginally outperforms R6-T and R7-T on all \mbox{R1--7} test sets. It slightly underperforms on the R5--7 test sets from \textsc{HatemojiBuild}.
This suggests that while training on a relatively small number of entries can substantially improve performance, the gains quickly saturate.
Future work could investigate performance differences in more detail using the labels for type and target of hate that we provide for all three rounds of data.

Due to practical constraints, \textsc{HatemojiCheck} has several limitations, which could be addressed in future work. First, it includes relatively simplistic statements, and so offers negative predictive power: high performance only indicates the absence of specific weaknesses \citep{gardnerEvaluatingModels2020a,rottgerHateCheckFunctional2021}.
Second, it is inspired by real-world hate but synthetically generated. Future work could evaluate against more complex and diverse forms of emoji-based hate.
Third, it is limited in scope. It only considers short English-language statements with one binary label. Only six identities are included, none of which are intersectional, and the set of emoji covers a fraction of the full Unicode Standard. In-scope identities and emoji usage are culturally-dependent \cite{barbieriHowCosmopolitan2016, ljubesicGlobalAnalysis2016} so future work could assess the generalizability of \textsc{HatemojiCheck} to other cultures and languages. These limitations do not diminish \textsc{HatemojiCheck}'s utility as a tool for effectively identifying model weaknesses to emoji-based hate. By publicly releasing the test suite, practitioners and academics can scrutinize their models prior to deployment or publication.

Adversarial data generation is a powerful technique for creating diverse, complex and informative datasets. 
However, it also introduces challenges.
First, the entries are `synthetic' rather than sampled from `real-world' examples. Substantial training and time are required to ensure that annotators understand real online hate and can imitate it. 
Second, annotators can exhaust their creativity and start producing unrealistic, simplistic or non-adversarial examples.
Third, because of the need for training, supervision and support, only a small pool of annotators is feasible, which can introduce additional idiosyncratic biases.
Given these issues, quality control is a key consideration throughout the data generation process. Encouragingly, we find carefully-curated and adversarially-generated training datasets can significantly improve performance on emoji-based hate as a particular type of challenging content, and that this approach is effective with relatively few training examples. Thus, substantial model improvements can be realized with minimal financial and computational cost.

\section{Related Work}

Emoji-based hate has received limited attention in prior work on hate and abusive language detection.
Studies that attend to emoji often do so as a potential input feature to aid classifier performance.
For example, \citet{samghabadiAttendingEmotions2019} improve offensive language classification with an `emotion-aware' mechanism built on emoji embeddings.
\citet{ibrohimIdentificationHate2019} find that adding emoji features to Continuous Bag of Words and word unigram models marginally improves performance for abusive language detection on Indonesian Twitter.
\citet{bickAnnotatingEmoticons2020} identifies examples of subtle and non-direct hate speech in German--Danish Twitter conveyed through `winking' or `skeptical' emoji that flag irony or non-literal meaning in their accompanying text.
\citet{corazzaHybridEmojiBased2020} train an emoji-based Masked Language Model (MLM) for zero-shot abuse detection. They show this method improves performance on classifying abuse in German, Italian and Spanish tweets compared to an MLM which does not attend to emoji. 
\citet{wiegandExploitingEmojis2021} use abusive emoji as a proxy for learning a lexicon of abusive words. Their findings indicate that emoji can disambiguate abusive and profane usages of words such as f*ck and b*tch.
By contrast, our work focuses on emoji-based hate as a challenge for hate detection models.
With \textsc{HatemojiCheck}, we enable a systematic evaluation of how well models handle different types of emoji-based hate. 
Rather than adjusting the model architecture, we account for emoji-based hate in our iterative data generation process and show that models trained on such data perform better on emoji-based hate, while retaining strong performance on text-only hate.

As a suite of functional tests for evaluation, \textsc{HatemojiCheck} directly builds on previous work by \citet{ribeiroAccuracyBehavioral2020} and \citet{rottgerHateCheckFunctional2021}.
\citet{ribeiroAccuracyBehavioral2020} introduced functional tests as a framework for NLP model evaluation with \textsc{CheckList}, showing that their approach can identify granular model strengths and weaknesses that are obscured by high-level metrics like accuracy and F1-score.
\citet{rottgerHateCheckFunctional2021} adapted this framework to hate detection with \textsc{HateCheck}, which covers 29 model functionalities motivated by interviews with civil society stakeholders and a review of previous hate speech literature.
Like \textsc{HateCheck}, we pair hateful test cases with contrasting perturbations that are particularly challenging to models relying on overly simplistic decision rules and thus reveal granular decision boundaries. \textsc{HateCheck} did not consider emoji-based hate, which is the main focus of our work.

Our approach to training better models for emoji-based hate builds directly on work by \citet{vidgenLearningWorst2021}, who apply iterative human-and-model-in-the-loop training to hate detection models.
Like us, they used the Dynabench interface \citep{kielaDynabenchRethinking2021} to implement their training system, which has also been used to improve model performance for other tasks such as reading
comprehension \citep{bartoloBeatAI2020} and sentiment analysis \citep{pottsDynaSentDynamic2021}.
Earlier work by \citet{dinanBuildIt2019} introduced a similar `build it, break it, fix it' system of repeated interactions between a hate classifier and crowdworkers to develop safe-by-design chatbots.
Unlike previous work, we focus data generation on a particular type of hateful content, emoji-based hate, and show that the training scheme can address specific model weaknesses on such content without sacrificing performance on text-only hate.

\section{Conclusion}
Online hate is a pervasive, harmful phenomenon, and hate detection models are a crucial tool for tackling it at scale.
We showed that emoji pose a particular challenge for such models, and presented \textsc{HatemojiCheck}, a first-of-its-kind evaluation suite for emoji-based hate. It covers seven functionalities with $3{,}930$ test cases. 
Using this test suite, we exposed clear weaknesses in the performance of commercial and academic models on emoji-based hate. To address these weaknesses, we created the \textsc{HatemojiBuild} dataset using an innovative human-and-model-in-the-loop approach. We showed that models trained on this adversarial data are substantially better at detecting emoji-based hate, while retaining strong performance on text-only hate. Our approach of first identifying granular model weaknesses, and then generating an adversarial dataset to address them presents a promising direction for building models to detect other diverse and emerging forms of online harm. 

\section*{Acknowledgments}
We are thankful for support that the Oxford authors received to facilitate annotation and computational resources from the Volkswagen Foundation, Meedan, Keble College, the Oxford Internet Institute, Rewire and The Alan Turing Institute.
We owe a debt of gratitude to all our annotators, to Dynabench and to our anonymous reviewers.
We are also grateful to Zeerak Talat and Douwe Kiela for their helpful advice on this research, as well as support from Devin Gaffney and Darius Kazemi.
Hannah Rose Kirk was supported by the Economic and Social Research Council grant ES/P000649/1.
Paul R\"ottger was supported by the German Academic Scholarship Foundation.

\bibliography{hatemoji_references}
\bibliographystyle{acl_natbib}

\clearpage
\appendix

\section{Ethical Considerations}
\paragraph{Misuse:} We release two datasets of challenging emoji examples on which commercial solutions and state-of-the-art transformer models have been proven to fail. Malicious actors could take inspiration for bypassing current detection systems on internet platforms, or in principal train a generative hate speech model. This concern is also raised by \citet{vidgenLearningWorst2021}, but the conclusion is reached that such risk of misuse is small and outweighed by the considerable scientific and social benefits. 

\paragraph{Harm Statement:} Following the advice of \citet{derczynskiHandlingPresenting2022a}, we describe the risks to well-being during the production and publication of this research and the steps taken to mitigate them. There is a risk of harm to data subjects i.e., the targets of hate, from reinforcing hateful, dehumanizing or derogatory statements, and to readers viewing this content in the paper. To mitigate these harms, we include a content warning directly after the abstract, at least a page before any harmful content is displayed and colored in red for maximum visibility. We also include a section-specific content warning before \S2.2, which includes a number of hateful examples. Hateful examples in \S2.2 are consistently formatted and visually distanced with gray text. All other examples (including \cref{tab:hmc_summary}, \cref{tab:func_hmc}) are presented with placeholders for the [IDENTITY]. Hateful slurs are starred out with an asterisk where possible (e.g., \cref{tab:slurs}) but we cannot star out emoji without hindering interpretation. There is also a risk to researchers and annotators from labeling and viewing harmful content. As authors, we oppose the use of hateful language. We follow protocols for protecting annotator well-being, including briefing sessions, regular check-ins and provision of mental health support.

\section{Data Statement for \textsc{HatemojiCheck}}
\label{appendix:A_data_statement}
We provide a data statement \citep{benderDataStatements2018} to document the generation and provenance of \textsc{HatemojiCheck}. 

\subsection{Curation Rationale}
To construct \textsc{HatemojiCheck}, we hand-crafted $3{,}930$ short-form English-language texts using a template-based method for group identities and slurs. Each test case exemplifies one functionality and is associated with a binary gold standard label (hate versus not hate). All $3{,}930$ cases were labeled by a trained team of three annotators, who could also flag examples that were unrealistic. Any test cases with multiple disagreements or flags were replaced with alternative templates and re-issued for annotation to improve the quality of examples in the final set of test cases. The purpose of \textsc{HatemojiCheck} is to evaluate the performance of black-box models against varied constructions of emoji-based hate.

\subsection{Language Variety}
The test cases are in English. This choice was motivated by the researchers' and annotators' expertise, and to maximize \textsc{HatemojiCheck}'s applicability to previous hate speech detection studies, which are predominantly conducted on English-language data. We discuss the limitations of restricting \textsc{HatemojiCheck} to one language and suggest that future work should prioritize expanding the test suite to other languages.

\subsection{Speaker Demographics}
All test cases were hand-crafted by the lead author, who is a native English-speaking researcher at a UK university with extensive subject matter expertise in online harms.

\subsection{Annotator Demographics}
We recruited a team of three annotators who worked for two weeks in May 2021 and were paid \textsterling16/hour. All annotators were female and between 30--39 years old. One had an undergraduate degree, one a taught graduate degree and one a post-graduate research degree. There were three nationalities: Argentinian, British and Iraqi, two ethnicities: White and Arab, and three religious affiliations: Catholic, Muslim and None. One annotator was a native English speaker and the others were non-native but fluent. All annotators used emoji and social media more than once per day. All annotators had seen others targeted by abuse online, and one had been targeted personally. 

\subsection{Speech Situation}
The modality of all test cases is written text embedded with emoji. The empirical investigation using Twitter's streaming API was conducted between 1st--14th April 2021. The first set of test cases was created between 26th April--7th May 2021. The first round of annotation ran between 7th--14th May 2021. The second round of cases was created and re-issued between 14th--21st May 2021.

\subsection{Text Characteristics}
The genre of texts is hateful and non-hateful statements using emoji constructions. Renderings of emoji vary by operating system and browser providers. The renderings in this paper are from WhatsApp. The composition of the dataset by labels (hate versus not hate) and by set (originals versus perturbations) is described in \cref{tab:agg_performance_hmc} of the main paper. 68\% of the test cases are hateful. \textsc{HatemojiCheck} has 644 cases for Muslims, 608 cases each for gay people, disabled people and women, 582 cases for Black people and 566 cases for trans people. identity perturbations switch a protected identity for a non-protected identity so 314 cases have no protected identity tag. 

\section{Constructing \textsc{HatemojiCheck}}
\label{appendix:A_constructing}
\subsection{List of Hateful Slurs in \textsc{HatemojiCheck}}
\label{appendix:A_slurs}
The slurs used in \textsc{HatemojiCheck} are selected in the same way as for \textsc{HateCheck} \cite{rottgerHateCheckFunctional2021}. For each of the six identities in \textsc{HatemojiCheck}, the most common slurs are taken from \url{hatebase.org}, an online crowd-sourced repository of hateful terms and slurs. The top three most common slurs logged by users of hatebase are used, unless the 3rd ranked slur is significantly less common than the 2nd ranked. Each identity thus has two or three associated slurs (\cref{tab:slurs}).

\subsection{Defining Perturbations}
\label{appendix:A_perturbations}
Each template has three perturbations but the mapping of original test cases to perturbations is not one-to-one. Instead, there is a set of original templates contrasted by three sets of perturbation templates. The perturbation templates have fewer degrees of freedom than the original templates so test cases generated from them are fewer in number. Within each perturbed template we use the same emoji for substitutions to ensure consistency, but across templates, we use a range of emoji to ensure linguistic variety.

The perturbed templates are constructed as follows. \textbf{(1) Identity perturbations}: the protected identity (which could be an emoji or a word depending on the functionality) in each original template is substituted with \textit{one} non-protected entity (which could be an emoji or a word). \textbf{(2) Polarity perturbations}: the negative term (which could be a word or an emoji) in each original template is substituted for \textit{one} positive term (which could be a word or an emoji). \textbf{(3) No emoji perturbations}: all emoji elements of the original template are replaced by equivalent text where \textit{one} word is used to cover all versions e.g. \emoji{gun} and \emoji{bomb} are both substituted for the verb `harm'. For F5 and F6, the appended emoji is removed not replaced.

\begin{table}[!t]
\centering
\footnotesize
\caption{Hateful slurs in \textsc{HatemojiCheck}}
\label{tab:slurs}
\begin{tabular}{ll} 
\toprule
\textbf{Protected Group} & \textbf{Slurs}               \\ 
\hline
Women                    & b*tch, sl*t, wh*re           \\
Trans people             & tr*nny, sh*male              \\
Gay people               & f*ggot, f*g, q*eer           \\
Black people             & n*gger, c*on                 \\
Disabled people          & r*tard, cr*pple, m*ng        \\
Muslims                  & m*zzie, J*hadi, camelf*cker  \\
\bottomrule
\end{tabular}
\end{table}

\section{Data Statement for \textsc{HatemojiBuild}}
\label{appendix:B_data_statement}
We provide a data statement \citep{benderDataStatements2018} to document the generation and provenance of \textsc{HatemojiBuild}. 

\subsection{Curation Rationale}
We use an online interface designed for dynamic dataset generation and model benchmarking (Dynabench) to collect synthetic adversarial examples in three successive rounds, running between 24th May--11th June. Each round contains ${\sim}$2,000 entries, where each original entry inputed to the interface is paired with an offline perturbation. Data was synthetically-generated by a team of trained annotators, i.e., not sampled from social media. 

\subsection{Language Variety}
All entries are in English. Language choice was dictated by the expertise of researchers and annotators. Furthermore, English is used for a wide number of benchmark hate speech datasets \citep{davidsonAutomatedHate2017, fountaLargeScale2018} and was also used in the adversarial dataset for textual hate speech \citep{vidgenLearningWorst2021}. The method be could be adapted for other languages in future work.

\subsection{Speaker Demographics}
All entries are synthetically-created by annotators so the speaker demographics match the annotator demographics.

\subsection{Annotator Demographics}
Ten annotators were recruited to work for three weeks, and paid \textsterling16/hour. An expert annotator was recruited for quality control purposes and paid \textsterling20/hour. In total, there were 11 annotators. All annotators received a training session prior to data collection and had previous experience working on hate speech projects. A daily `stand-up' meeting was held every morning to communicate feedback and update guidelines as rounds progressed. Annotators were able to contact the research team at any point using a messaging platform. Of 11 annotators, 8 were between 18--29 years old and 3 between 30--39 years old. The completed education level was high school for 3 annotators, undergraduate degree for 1 annotator, taught graduate degree for 4 annotators and post-graduate research degree for 3 annotators. 6 annotators were female, and 5 were male. Annotators came from a variety of nationalities, with 7 British, as well as Jordanian, Irish, Polish and Spanish. 7 annotators identified as ethnically White and the remaining annotators came from various ethnicities including Turkish, Middle Eastern, and Mixed White and South Asian. 4 annotators were Muslim, and others identified as Atheist or as having no religious affiliation. 9 annotators were native English speakers and 2 were non-native but fluent. The majority of annotators (9) used emoji and social media more than once per day. 10 annotators had seen others targeted by abuse online, and 7 had been personally targeted. 

\subsection{Speech Situation}
Entries were created from 24th May--11th June 2021. Their modality is short-form written texts embedded with emoji. Entries are synthetically-generated but annotators were trained on real-world examples of emoji-based hate from Twitter. 

\subsection{Text Characteristics}
The genre of texts is hateful and non-hateful statements using emoji constructions. Annotators inputted emoji into the platform using a custom emoji picker.\footnote{\url{https://hatemoji.stackblitz.io/}} The composition of the final dataset is described in \cref{tab:appendix_round_summary_stats}. 50\% of the $5{,}912$ test cases are hateful. 50\% of the entries in the dataset are original content and 50\% are perturbations. 

\begin{table}
\centering
\footnotesize
\setlength{\extrarowheight}{0pt}
\addtolength{\extrarowheight}{\aboverulesep}
\addtolength{\extrarowheight}{\belowrulesep}
\setlength{\aboverulesep}{0pt}
\setlength{\belowrulesep}{0pt}
\caption{Summary statistics across three rounds of data from \textsc{HatemojiBuild}.}
\label{tab:appendix_round_summary_stats}
\setlength{\tabcolsep}{1pt}
\resizebox{\columnwidth}{!}{%
\begin{tabular}{llccc} 
\toprule
                                                 &                                                     & \textbf{R5 }                                       & \textbf{R6 }                                       & \textbf{R7 }                                        \\ 
\hline
\textbf{$\bm{n}$}                                     &                                                     & 1994                                               & 1966                                               & 1952                                                \\ 
\hline
\multirow{3}{*}{\textbf{Split, $\bm{n}$ (\%) }}         & {\cellcolor[rgb]{0.753,0.753,0.753}}Train           & {\cellcolor[rgb]{0.753,0.753,0.753}}1595 (80.0)    & {\cellcolor[rgb]{0.753,0.753,0.753}}1572 (80.0)    & {\cellcolor[rgb]{0.753,0.753,0.753}}1561 (80.0)     \\
                                                 & Dev                                                 & 199 (10.0)                                         & 197 (10.0)                                         & 195 (10.0)                                          \\
                                                 & {\cellcolor[rgb]{0.753,0.753,0.753}}Test            & {\cellcolor[rgb]{0.753,0.753,0.753}}200 (10.0)     & {\cellcolor[rgb]{0.753,0.753,0.753}}197 (10.0)     & {\cellcolor[rgb]{0.753,0.753,0.753}}196 (10.0)      \\ 
\hline
\multirow{2}{*}{\textbf{Label, $\bm{n}$ (\%)~}}         & Hate                                                & 1006 (50.5)                                        & 983 (50.0)                                         & 976 (50.0)                                          \\
                                                 & {\cellcolor[rgb]{0.753,0.753,0.753}}Not hate        & {\cellcolor[rgb]{0.753,0.753,0.753}}988 (49.5)     & {\cellcolor[rgb]{0.753,0.753,0.753}}983 (50.0)     & {\cellcolor[rgb]{0.753,0.753,0.753}}976 (50.0)      \\ 
\hline
\multirow{2}{*}{\textbf{\textbf{Set, $\bm{n}$ (\%)}}} & Original                                            & 997 (50.0)                                         & 983 (50.0)                                         & 976 (50.0)                                          \\
                                                 & {\cellcolor[rgb]{0.753,0.753,0.753}}Perturbation    & {\cellcolor[rgb]{0.753,0.753,0.753}}997 (50.0)     & {\cellcolor[rgb]{0.753,0.753,0.753}}983 (50.0)     & {\cellcolor[rgb]{0.753,0.753,0.753}}976 (50.0)      \\ 
\hline
\multirow{5}{*}{\textbf{\textbf{Type, $\bm{n}$ (\%)}}}  & None                                                & 988 (49.5)                                         & 983 (50.0)                                         & 976 (50.0)                                          \\
                                                 & {\cellcolor[rgb]{0.753,0.753,0.753}}Derogation      & {\cellcolor[rgb]{0.753,0.753,0.753}}718 (36.0)     & {\cellcolor[rgb]{0.753,0.753,0.753}}649 (33.0)     & {\cellcolor[rgb]{0.753,0.753,0.753}}594 (30.4)      \\
                                                 & Animosity                                           & 74 (3.7)                                           & 219 (11.1)                                         & 275 (14.1)                                          \\
                                                 & {\cellcolor[rgb]{0.753,0.753,0.753}}Threatening & {\cellcolor[rgb]{0.753,0.753,0.753}}101 (5.1)      & {\cellcolor[rgb]{0.753,0.753,0.753}}50 (2.5)       & {\cellcolor[rgb]{0.753,0.753,0.753}}52 (2.7)        \\
                                                 & Dehumanizing                                    & 113 (5.7)                                          & 65 (3.3)                                           & 55 (2.8)                                            \\ 
\hline
\textbf{\# Emoji, $\mu$ ($\sigma$) }           & {\cellcolor[rgb]{0.753,0.753,0.753}}                & {\cellcolor[rgb]{0.753,0.753,0.753}}1.7 (2.2) & {\cellcolor[rgb]{0.753,0.753,0.753}}1.7 (1.0) & {\cellcolor[rgb]{0.753,0.753,0.753}}1.6 (1.1) \\
\bottomrule
\end{tabular}
}
\end{table}

\section{Constructing \textsc{HatemojiBuild}}

\label{appendix:B_composition}

\paragraph{Types of Hate}
We adopt the same categorization used by \citet[p.3]{vidgenLearningWorst2021}.\footnote{One of their categories ``support for hateful entities'' is excluded because it introduced confusion and ambiguity.} There are four types of hate.
Derogation: Language which explicitly derogates, demonizes, demeans or insults a group. 
Animosity: Expressions of abuse through implicit statements or mockery, where a logical step must be taken between the sentence and its intended negativity.
Threatening language: Statements of intent to take action against a group, with the potential to inflict serious or imminent harm on its members. 
Dehumanizing language: Comparing groups to insects, animals, germs or trash. 

\paragraph{Targets of Hate} Annotators were provided with a non-exhaustive list of high-priority identities to focus on which included categorizations by gender identity (e.g., women, trans), sexual orientation (e.g., gay), ethnicity (e.g., Hispanic people), religion (e.g., Sikh), nationality (e.g., Polish), disability and class, alongside intersections (e.g., Muslim women). Hate directed towards majority groups (e.g., men, white people and heterosexuals) is outside the remit of this work. The explicit decision not to focus on issues such as `reverse racism' \citep{baxCWordMeets2018} is made due to the complex debate on its inclusion in hate speech definitions.

\paragraph{Composition} \cref{tab:cases_flow} reports the inter-annotator agreement and number of final entries per round. Each round has approximately the same number of entries with slightly fewer in R7 due to more quality control issues. Labels are equally distributed in each round (see \cref{tab:appendix_round_summary_stats}). For 75 pairs of originals and perturbations, the perturbation unsuccessfully flipped the label given by majority agreement between three annotators. All other pairs have opposite labels. Derogation is always the most-commonly inputted form of hate. From R5 to R7, there is a rise in animosity entries paired with a decline in threatening and dehumanizing language entries. Annotators were given substantial freedom in the targets of hate resulting in $54$ unique targets, and $126$ unique intersections of these. The entries from R5--R7 contain $1{,}082$ unique emoji out of $3{,}521$ defined in the Unicode Standard as of September 2020. The mode of emoji per entries is $1$ and mean is approximately $1.5$ in each round. The frequency of targets and emoji follow a long-tailed distribution, similar to a Zipf curve. These distributions match those found online for targets \citep{silvaAnalyzingTargets2016} and for emoji \citep{felboUsingMillions2017, cappalloNewModality2019, bickAnnotatingEmoticons2020}. 

In the first round, annotators commonly employed a strategy of substituting identities for emoji so three identity emoji were frequently used (\emoji{black}, \emoji{male_couple}, \emoji{dis}). Using emoji character substitutions in slurs was also a successful strategy (A: \emoji{A}), as was substituting the entire slur for an emoji homonym (f*g: \emoji{cig}). In the final round, the model was wise to such strategies, and so annotators changed their adversarial techniques to irony, satire or mockery, as reflected in the top emoji (\emoji{tears_of_joy}, \emoji{rolling_eyes}, \emoji{rofl}, \emoji{clown}). For a qualitative understanding of annotators' experiences, we conducted a post-study survey.\footnote{This survey is on our \href{https://github.com/HannahKirk/Hatemoji/blob/main/ExtraResources/HatemojiBuild_Annotator_Survey.pdf}{Github}.}

\begin{table}[!t]
\centering
\footnotesize
\caption{Quality control checks per round of data collection, showing Randolph-kappa score across three annotators per entry $R(S)$, entries reviewed by an expert annotator, and dropped entries.}
\label{tab:cases_flow}
\setlength{\extrarowheight}{0pt}
\addtolength{\extrarowheight}{\aboverulesep}
\addtolength{\extrarowheight}{\belowrulesep}
\setlength{\aboverulesep}{0pt}
\setlength{\belowrulesep}{0pt}
\setlength{\tabcolsep}{1pt}
\resizebox{\columnwidth}{!}{%
\begin{tabular}{llccccc} 
\toprule
                             &                                              & \textbf{Input Entries}                   & \textbf{$\bm{R(S)}$}                             & \textbf{Reviewed}                       & \textbf{Dropped}                       & \textbf{Final Entries}                    \\ 
\hline
\multirow{2}{*}{\textbf{R5}} & {\cellcolor[rgb]{0.753,0.753,0.753}}Original & {\cellcolor[rgb]{0.753,0.753,0.753}}1020 & {\cellcolor[rgb]{0.753,0.753,0.753}}0.903 & {\cellcolor[rgb]{0.753,0.753,0.753}}273 & {\cellcolor[rgb]{0.753,0.753,0.753}}6  & {\cellcolor[rgb]{0.753,0.753,0.753}}1014  \\
                             & Perturbation                                 & 1014                                     & 0.938                                     & 131                                     & 17                                     & \textbf{997}                              \\ 
\hline
\multirow{2}{*}{\textbf{R6}} & {\cellcolor[rgb]{0.753,0.753,0.753}}Original & {\cellcolor[rgb]{0.753,0.753,0.753}}1008 & {\cellcolor[rgb]{0.753,0.753,0.753}}0.902 & {\cellcolor[rgb]{0.753,0.753,0.753}}189 & {\cellcolor[rgb]{0.753,0.753,0.753}}8  & {\cellcolor[rgb]{0.753,0.753,0.753}}1000  \\
                             & Perturbation                                 & 1000                                     & 0.911                                     & 167                                     & 17                                     & \textbf{983}                              \\ 
\hline
\multirow{2}{*}{\textbf{R7}} & {\cellcolor[rgb]{0.753,0.753,0.753}}Original & {\cellcolor[rgb]{0.753,0.753,0.753}}993  & {\cellcolor[rgb]{0.753,0.753,0.753}}0.913 & {\cellcolor[rgb]{0.753,0.753,0.753}}169 & {\cellcolor[rgb]{0.753,0.753,0.753}}12 & {\cellcolor[rgb]{0.753,0.753,0.753}}981   \\
                             & Perturbation                                 & 981                                      & 0.928                                     & 168                                     & 5                                      & \textbf{976}                              \\
\bottomrule
\end{tabular}
}
\end{table}

\begin{table*}[t]
\centering
\footnotesize
\caption{Performance of models on emoji, text and all adversarial test sets, alongside benchmark evaluation sets: \textsc{HatemojiCheck} and \textsc{HateCheck}.}
\label{tab:appendix_agg_performance_db}
\setlength{\aboverulesep}{0pt}
\setlength{\belowrulesep}{0pt}
\setlength{\tabcolsep}{6pt}
\begin{tabular}{lcccc|cccc|cc} 
\toprule
                & \multicolumn{4}{c|}{\textbf{Emoji Test Sets}}                                                                                                                                                                     & \multicolumn{4}{c|}{\textbf{Text Test Sets}}                                                                                                                                                                      & \multicolumn{2}{c}{\textbf{All Rounds}}                                                                  \\ 
\cline{2-11}
                & \multicolumn{2}{c}{\textbf{R5-R7}}                                                                      & \multicolumn{2}{c|}{\textbf{\textsc{HmojiCheck}}}                                                                & \multicolumn{2}{c}{\textbf{R1-R4}}                                                                      & \multicolumn{2}{c|}{\textbf{\textsc{HateCheck}}}                                                                 & \multicolumn{2}{c}{\textbf{R1-R7}}                                                                       \\ 
               & \multicolumn{2}{c}{\textbf{$n=593$}}                                                                      & \multicolumn{2}{c|}{\textbf{$n=3930$}}                                                                & \multicolumn{2}{c}{\textbf{$n=4119$}}                                                                      & \multicolumn{2}{c|}{\textbf{$n=3728$}}                                                                 & \multicolumn{2}{c}{\textbf{$n=4712$}}                                                                       \\ 
\cline{2-11}
                & \multicolumn{1}{c}{\textbf{Acc}}                   & \multicolumn{1}{c}{\textbf{F1}}                    & \multicolumn{1}{c}{\textbf{Acc}}                   & \multicolumn{1}{c|}{\textbf{F1}}                   & \multicolumn{1}{c}{\textbf{Acc}}                   & \multicolumn{1}{c}{\textbf{F1}}                    & \multicolumn{1}{c}{\textbf{Acc}}                   & \multicolumn{1}{c|}{\textbf{F1}}                   & \multicolumn{1}{c}{\textbf{Acc}}                   & \multicolumn{1}{c}{\textbf{F1}}                     \\ 
\hline
\textbf{P-IA}  & {\cellcolor[rgb]{0.929,0.631,0.604}}0.508          & {\cellcolor[rgb]{0.961,0.804,0.792}}0.394          & {\cellcolor[rgb]{0.973,0.871,0.863}}0.689          & {\cellcolor[rgb]{0.98,0.898,0.894}}0.754           & {\cellcolor[rgb]{0.961,0.804,0.788}}0.679          & {\cellcolor[rgb]{0.973,0.859,0.847}}0.720          & {\cellcolor[rgb]{0.961,0.808,0.796}}0.765          & {\cellcolor[rgb]{0.969,0.843,0.831}}0.839          & {\cellcolor[rgb]{0.961,0.8,0.784}}0.658            & {\cellcolor[rgb]{0.973,0.863,0.855}}0.689           \\
\textbf{P-TX} & {\cellcolor[rgb]{0.949,0.753,0.733}}0.523          & {\cellcolor[rgb]{0.988,0.945,0.941}}0.448          & {\cellcolor[rgb]{0.953,0.761,0.745}}0.650          & {\cellcolor[rgb]{0.957,0.78,0.765}}0.711           & {\cellcolor[rgb]{0.922,0.592,0.565}}0.602          & {\cellcolor[rgb]{0.945,0.714,0.694}}0.659          & {\cellcolor[rgb]{0.945,0.718,0.698}}0.720          & {\cellcolor[rgb]{0.957,0.776,0.761}}0.813          & {\cellcolor[rgb]{0.922,0.592,0.565}}0.592          & {\cellcolor[rgb]{0.949,0.741,0.725}}0.639           \\
\textbf{B-D}    & {\cellcolor[rgb]{0.902,0.486,0.451}}0.489          & {\cellcolor[rgb]{0.902,0.486,0.451}}0.270          & {\cellcolor[rgb]{0.914,0.557,0.529}}0.578          & {\cellcolor[rgb]{0.918,0.573,0.541}}0.636          & {\cellcolor[rgb]{0.914,0.557,0.529}}0.589          & {\cellcolor[rgb]{0.922,0.588,0.561}}0.607          & {\cellcolor[rgb]{0.91,0.545,0.514}}0.632           & {\cellcolor[rgb]{0.922,0.592,0.565}}0.738          & {\cellcolor[rgb]{0.922,0.588,0.561}}0.591          & {\cellcolor[rgb]{0.925,0.616,0.588}}0.586           \\
\textbf{B-F}    & {\cellcolor[rgb]{0.91,0.537,0.506}}0.496           & {\cellcolor[rgb]{0.925,0.62,0.592}}0.322           & {\cellcolor[rgb]{0.902,0.486,0.451}}0.552          & {\cellcolor[rgb]{0.902,0.486,0.451}}0.605          & {\cellcolor[rgb]{0.902,0.486,0.451}}0.562          & {\cellcolor[rgb]{0.902,0.486,0.451}}0.562          & {\cellcolor[rgb]{0.902,0.486,0.451}}0.602          & {\cellcolor[rgb]{0.902,0.486,0.451}}0.694          & {\cellcolor[rgb]{0.902,0.486,0.451}}0.557          & {\cellcolor[rgb]{0.902,0.486,0.451}}0.532           \\
\textbf{R5-T}   & {\cellcolor[rgb]{0.902,0.961,0.933}}0.585          & {\cellcolor[rgb]{0.957,0.984,0.969}}0.490          & {\cellcolor[rgb]{0.796,0.918,0.859}}0.779          & {\cellcolor[rgb]{0.808,0.925,0.867}}0.825          & {\cellcolor[rgb]{0.341,0.733,0.541}}\textbf{0.828} & {\cellcolor[rgb]{0.341,0.733,0.541}}\textbf{0.847} & {\cellcolor[rgb]{0.4,0.761,0.584}}0.956            & {\cellcolor[rgb]{0.404,0.761,0.588}}0.968          & {\cellcolor[rgb]{0.541,0.816,0.682}}0.786          & {\cellcolor[rgb]{0.565,0.824,0.698}}0.801           \\
\hline
\textbf{R6-T}   & {\cellcolor[rgb]{0.349,0.737,0.545}}0.757          & {\cellcolor[rgb]{0.341,0.733,0.541}}\textbf{0.769} & {\cellcolor[rgb]{0.341,0.733,0.541}}\textbf{0.879} & {\cellcolor[rgb]{0.341,0.733,0.541}}\textbf{0.910} & {\cellcolor[rgb]{0.384,0.753,0.573}}0.823          & {\cellcolor[rgb]{0.439,0.773,0.608}}0.837          & {\cellcolor[rgb]{0.376,0.749,0.565}}0.961          & {\cellcolor[rgb]{0.38,0.749,0.569}}0.971           & {\cellcolor[rgb]{0.349,0.737,0.549}}0.813          & {\cellcolor[rgb]{0.373,0.749,0.565}}0.825           \\
\textbf{R7-T}   & {\cellcolor[rgb]{0.341,0.733,0.541}}\textbf{0.759} & {\cellcolor[rgb]{0.361,0.741,0.553}}0.762          & {\cellcolor[rgb]{0.4,0.757,0.58}}0.867             & {\cellcolor[rgb]{0.404,0.761,0.584}}0.899          & {\cellcolor[rgb]{0.376,0.749,0.565}}0.824          & {\cellcolor[rgb]{0.388,0.753,0.573}}0.842          & {\cellcolor[rgb]{0.408,0.761,0.588}}0.955          & {\cellcolor[rgb]{0.412,0.765,0.592}}0.967          & {\cellcolor[rgb]{0.345,0.737,0.545}}0.813          & {\cellcolor[rgb]{0.345,0.737,0.545}}0.829           \\
\textbf{R8-T}   & {\cellcolor[rgb]{0.392,0.757,0.576}}0.744          & {\cellcolor[rgb]{0.376,0.749,0.565}}0.755          & {\cellcolor[rgb]{0.38,0.749,0.569}}0.871           & {\cellcolor[rgb]{0.376,0.749,0.569}}0.904          & {\cellcolor[rgb]{0.349,0.737,0.545}}0.827          & {\cellcolor[rgb]{0.373,0.749,0.565}}0.844          & {\cellcolor[rgb]{0.341,0.733,0.541}}\textbf{0.966} & {\cellcolor[rgb]{0.341,0.733,0.541}}\textbf{0.975} & {\cellcolor[rgb]{0.341,0.733,0.541}}\textbf{0.814} & {\cellcolor[rgb]{0.341,0.733,0.541}}\textbf{0.829}  \\ 
\bottomrule
\end{tabular}
\setlength{\aboverulesep}{\origaboverulesep} %
\end{table*}
\section{Target Models}
\label{appendix:B_model_training}
In each round we assessed two candidate model architectures. The first is an uncased DeBERTa base model (134M params) with a sequence classification head \citep{heDeBERTaDecodingenhanced2021}. DeBERTa has been shown to improve on BERT \citep{devlinBERTPretraining2019} and RoBERTa \citep{liuRoBERTaRobustly2019} models by incorporating a disentangled attention mechanism and enhanced mask decoder. However, emoji are likely relatively sparse in DeBERTa's pre-training material which includes English Wikipedia, Book Corpus \citep{zhuAligningBooks2015} and a subset of CommonCrawl. DeBERTa uses the BPE-encoding method so the tokenizer can encode emoji as unique tokens. The second candidate is an uncased BERTweet model (135M params) with a sequence classification head where the RoBERTa training procedure was repeated on 850M English Tweets using a custom vocabulary \citep{nguyenBERTweetPretrained2020}. BERTweet tokenizes emoji by first translating them to text representations using the \texttt{emoji} package. These word representations are derived from the Unicode Standard, so for example \emoji{crying_face} becomes \texttt{:crying\_face:}.
 
Upsampling ratios are evaluated for each new round of training data with multipliers of $1,5,10,100$. For the R1--4 data, we carry forward the upsampling from \citet{vidgenLearningWorst2021}: R1 is upsampled five times, R2 is upsampled 100 times, R3 is upsampled once, and R4 is upsampled once. Combining the model architectures and upsampling ratios gives 8 candidate models for each round's target model. All models were implemented using the \texttt{transformers} library \citep{wolfHuggingFaceTransformers2020}. All models were trained for 3 epochs with early stopping based on the dev set loss, a learning rate of $2e-5$ and a weighted Adam optimizer. Other hyperparameters were set to HuggingFace defaults. We train and evaluate each model once and report results for this single run. Training took approximately 7 hours for each BERTweet model and 15 hours for each DeBERTa model using 8 NVIDIA Tesla V100 GPU on the JADE2 supercomputer.

The best target model for each round is selected by weighted accuracy between all prior rounds and the current round. For R6-T, a DeBERTa model with 100x upsampling on R5 performs best. Given the dearth of emoji in R1--R4, it is unsurprising a large upsample improves performance on emoji-based hate.
For R7-T, DeBERTa with one upsample on R6 performs best.
For R8-T, DeBERTa with five upsamples on R7 performs best.
In all rounds, DeBERTa significantly outperforms BERTweet, while the upsampling ratio less substantially affects performance. This suggests first converting emoji to their text representations does not substantially aid performance, though there may be other differences between the models driving results. We  upload models to Dynalab for model-in-the-loop evaluation and data collection \citep{maDynaboardEvaluationAsAService2021}.

\section{Robustness Analysis of Baselines}

\label{appendix:B_baselines}
In addition to the models analyzed in the main paper, we evaluate three further models. The first is the `toxicity' attribute returned by Perspective API (P-TX). We test this model because toxicity ratings are the most popular attributes.\footnote{\url{https://support.perspectiveapi.com/s/about-the-api-faqs}} The second and third models are two uncased BERT models \citep{devlinBERTPretraining2019} trained on publicly-available academic datasets. B-D is trained on the \citet{davidsonAutomatedHate2017} dataset of $24{,}783$ tweets, labeled as \textit{hateful}, \textit{offensive} and \textit{neither}. B-F is trained on the \citet{fountaLargeScale2018} dataset of $99{,}996$ tweets, labeled as \textit{hateful}, \textit{abusive}, \textit{spam} and \textit{normal}. Any labels besides \textit{hateful} are binarized into a single \textit{non-hateful} label. Two factors motivated the testing of these models. (1) There is a lack of emoji in the training data: the \citeauthor{davidsonAutomatedHate2017} dataset has 5.8\% hateful cases, but only 7.4\% of these contain emoji, and the \citeauthor{fountaLargeScale2018} dataset has 5.0\% hateful cases, of which 14.7\% contain emoji. (2) Despite BERT being a commonly-used architecture for hate speech detection, it encodes emoji as \texttt{<UNK>} tokens by default. We compare performance of the full set of pre-emoji models including those analyzed in the main paper (P-IA, P-TX, B-D, B-F, R5-T) versus our `emoji-aware' models from each round of data collection (R6-T, R7-T, R8-T). \cref{tab:appendix_agg_performance_db} shows performance against the adversarially produced datasets in the emoji rounds of \textsc{HatemojiBuild} and text rounds from \citet{vidgenLearningWorst2021}, alongside two benchmark evaluation sets \textsc{HatemojiCheck} and \textsc{HateCheck} \cite{rottgerHateCheckFunctional2021}. P-TX has comparable performance to P-IA. B-D and B-F perform poorly on \textsc{HatemojiCheck} (F1 = 0.64, 0.61), and even more poorly on the adversarial test sets of emoji (F1 = 0.27, 0.32).

\section{Fairness Considerations}
\label{appendix:B_fairness}
As well as the six metrics discussed in the main paper, \cref{tab:fairness_ratios} shows two further fairness metrics: (1) the demographic parity ratio \citep{agarwalFairRegression2019}, which is equal to one when the selection rate is balanced across subgroups, and (2) the equalized odds ratio \citep{hardtEqualityOpportunity2016, agarwalReductionsApproach2018a}, which is equal to one when the true positive, true negative, false positive, and false negative rates are balanced across subgroups. The academic BERT models perform worst (0.14 to 0.28 across the metrics), followed by the Perspective models (0.41 to 0.67). The most fair model is {R5-T with Demographic Parity Ratio of 0.90 and Equalized Odds Ratio of 0.77. However, the overall performance of the emoji-aware models is higher (see \cref{fig:fairness}): for every subgroup, R8-T has higher accuracy, precision and recall, and lower false positive and negative rates. Thus, the marginally worse performance by between-group fairness metrics is paired with a better ability to protect all these subgroups against emoji-based hate.

\begin{table}[H]
\centering
\footnotesize
\caption{Between-group fairness metrics by model.}
\label{tab:fairness_ratios}
\setlength{\aboverulesep}{0pt}
\setlength{\belowrulesep}{0pt}
\setlength{\tabcolsep}{3pt}
\begin{tabular}{lcc} 
\toprule
              & \textbf{Demographic}                     & \textbf{Equalized}                        \\
              & \textbf{Parity Ratio}                    & \textbf{Odds Ratio}                       \\ 
\hline
\textbf{P-IA} & {\cellcolor[rgb]{0.984,0.929,0.925}}0.668 & {\cellcolor[rgb]{0.976,0.882,0.875}}0.430  \\
\textbf{P-TX} & {\cellcolor[rgb]{0.976,0.878,0.871}}0.611 & {\cellcolor[rgb]{0.969,0.839,0.831}}0.405  \\
\textbf{B-D}  & {\cellcolor[rgb]{0.902,0.486,0.451}}0.174 & {\cellcolor[rgb]{0.902,0.486,0.451}}0.138  \\
\textbf{B-F}  & {\cellcolor[rgb]{0.918,0.584,0.553}}0.276 & {\cellcolor[rgb]{0.933,0.647,0.624}}0.257  \\
\textbf{R5-T} & {\cellcolor[rgb]{0.341,0.733,0.541}}\textbf{0.898} & {\cellcolor[rgb]{0.341,0.733,0.541}}\textbf{0.767}  \\
\hline
\textbf{R6-T} & {\cellcolor[rgb]{0.682,0.875,0.78}}0.818  & {\cellcolor[rgb]{0.498,0.796,0.651}}0.711 \\
\textbf{R7-T} & {\cellcolor[rgb]{0.557,0.82,0.69}}0.848   & {\cellcolor[rgb]{0.6,0.839,0.722}}0.667    \\
\textbf{R8-T} & {\cellcolor[rgb]{0.639,0.855,0.749}}0.832 & {\cellcolor[rgb]{0.784,0.914,0.851}}0.600  \\ 
\bottomrule
\end{tabular}
\end{table}

\end{document}